\documentclass[10pt,a4paper]{article}
\usepackage[top=1in, bottom=1.25in, left=1in, right=1in]{geometry}
\usepackage[utf8x]{inputenc}
\usepackage{times}
\usepackage{fancyhdr}
\usepackage{fancybox}
\usepackage{changepage}
\usepackage{xcolor}

\usepackage[round,comma]{natbib} 
\usepackage{appendix} 

\usepackage{algorithm,algorithmic} 

\usepackage{amsmath,amsfonts,amssymb}
\usepackage{mathletters} 

\newcommand{\beq}{\begin{equation}}
\newcommand{\eeq}{\end{equation}}
\newcommand{\beqa}{\begin{eqnarray}}
\newcommand{\eeqa}{\end{eqnarray}}
\newcommand{\beqan}{\begin{eqnarray*}}
\newcommand{\eeqan}{\end{eqnarray*}}

\newtheorem{theorem}{Theorem}
\newtheorem{lemma}{Lemma}
\newtheorem{corollary}{Corollary}
\newtheorem{definition}{Definition}
\newtheorem{assumption}{Assumption}
\newtheorem{remark}{Remark}

\newcommand{\argmax}{\mathop{\mathrm{argmax}}}

\newcommand{\Argmax}{\mathop{\mathrm{Argmax}}}

\newcommand{\meas}{\kM_1(\Real)}

\newcommand{\KLUCB}{\textcolor{red!60!black}{\texttt{KL-ucb}}}
\newcommand{\KLUCBp}{\textcolor{red!60!black}{\texttt{KL-ucb+}}}
\renewcommand{\KL}{\texttt{KL}}
\newcommand{\norm}[1]{\|#1\|}
\newcommand{\bX}{\mathbb{X}}

\newenvironment{myproof}[1]{

	\begin{adjustwidth}{0.1cm}{0.1cm}	
	
			\noindent
			\hrulefill
			
				\noindent{\bf Proof #1: }				
	}
	{$\hfill\square$
	
		\vspace{-2mm}
	\noindent
	\hrulefill
\end{adjustwidth}
\bigskip
	}

\newenvironment{mytheorem}[1]{

	\begin{minipage}{0.91\textwidth}
		\vspace{2mm}	
		\begin{theorem}[#1]
}{
\end{theorem}
\vspace{0mm}
\end{minipage}

}

\newcommand{\theorembox}[3]{
\bigskip\framebox{\begin{mytheorem}{#1}\label{thm:#2}#3
\end{mytheorem}}	
}

\newenvironment{mycorollary}[1]{
	
	\begin{minipage}{0.91\textwidth}
		\vspace{2mm}	
		\begin{corollary}[#1]
		}{
		\end{corollary}
		\vspace{0mm}
	\end{minipage}
	
}

\newcommand{\corollarybox}[3]{
	\bigskip\framebox{\begin{mycorollary}{#1}\label{cor:#2}#3
	\end{mycorollary}}	
}
			
\newenvironment{mylemma}[1]{

		\begin{minipage}{0.91\textwidth}
			\vspace{2mm}	
			\begin{lemma}[#1]
	}{
\end{lemma}
\vspace{0mm}
\end{minipage}

	}

\newcommand{\lemmabox}[3]{
	\bigskip\framebox{\begin{mylemma}{#1}\label{lem:#2}#3
	\end{mylemma}}	
	}	
	
\newcommand{\andAt}{\ \,\, \mbox{\small and} \ \,\, a_{t+1} = a}

\begin{document}

\thispagestyle{empty}
\begin{center}
{\Large {\bf Boundary Crossing Probabilities for General Exponential Families}}

\medskip
	\textsc{Odalric-Ambrym Maillard}\\
	\textit{INRIA Lille - Nord Europe}\\
	\textit{40 Avenue Halley}\\ \textit{59650 Villeneuve d'Ascq, France}\\
	\texttt{odalricambrym.maillard@inria.fr}\\
\end{center}

We consider parametric exponential families of dimension $K$ on the real line.
We study a variant of \textit{boundary crossing probabilities} coming from 
the multi-armed bandit literature, in the case when the real-valued distributions form an exponential family of dimension $K$. Formally, our result is a concentration inequality that bounds the probability that $\cB^\psi(\hat \theta_n,\theta^\star)\geq f(t/n)/n$, where $\theta^\star$ is the parameter of an unknown target distribution,
$\hat \theta_n$ is the empirical parameter estimate built from $n$ observations, $\psi$ is the log-partition function of the exponential family and $\cB^\psi$ is the corresponding Bregman divergence. From the perspective of stochastic multi-armed bandits, we pay special attention to the case when the boundary function $f$ is logarithmic, as it is enables to analyze the regret of the state-of-the-art \KLUCB\ and \KLUCBp\ strategies, whose analysis was left open in such generality. Indeed, previous results only hold for the case when $K=1$,
while we provide results for arbitrary finite dimension $K$, thus considerably extending the existing results.
Perhaps surprisingly, we highlight that the proof techniques to achieve these strong results  already  existed three decades ago in the work of T.L. Lai, and were apparently forgotten in the bandit community. We provide a modern rewriting of these beautiful techniques that we believe are useful beyond the application to stochastic multi-armed bandits. 


\section{Multi-armed bandit setup and notations}\label{sec:notations}
Let us consider a stochastic multi-armed bandit problem $(\cA,\nu)$, where $\cA$ is a finite set
of cardinality $A\in\Nat$ and $\nu = (\nu_a)_{a\in\cA}$ is a set of probability distribution over $\Real$
indexed by $\cA$.  The game is sequential and goes as follows:
\begin{center}
At each round $t \in\Nat$, the player picks an arm $a_t$ (based on her past observations)
and receives a stochastic payoff $Y_t$ drawn independently at random according to the distribution $\nu_{a_t}$. She only observes the payoff $Y_t$, and her goal is to maximize her expected cumulated payoff, $\sum_{t=1} Y_{a_t}$, over a possibly unknown number of steps.
\end{center}

Although the term multi-armed bandit problem was probably coined during the 60's in reference to the casino slot machines of the 19th century, the formulation of this problem is due to Herbert Robbins -- one of the most brilliant mind of his time, see \cite{robbins1952some} and takes its origin in earlier questions about optimal stopping policies for clinical trials, see \cite{thompson1933likelihood,thompson1935criterion,wald1945sequential}. We refer the interested reader to \cite{robbins2012herbert} regarding
the legacy of the immense work of H.~Robbins in mathematical statistics for the sequential design of experiments, compiling his most outstanding research for his 70's birthday.
Since then, the field of multi-armed bandits has grown large and bold, and we humbly refer to the introduction of \cite{CaGaMaMuSt2013}
for key historical aspects about the development of the field. 
Most notably, they include first the introduction of dynamic allocation indices (aka Gittins indices,  \cite{Gittins79bandits}) 
suggesting that an optimal strategy can be found in the form of an index strategy (that at each round selects an arm with highest "index"); second, the seminal work of \cite{LaiRobbins85bandits} that shows indexes can be chosen as "upper confidence bounds" on the mean reward of each arm, and provided the first asymptotic lower-bound on the achievable performance for specific distributions;
third, the generalization of this lower bound in the 90's to generic distributions by \cite{BurnetasKatehakis97bandits} (see also the recent work from \cite{garivier2016explore}) as well as the 
asymptotic analysis by \cite{agrawal1995sample} of generic classes of upper-confidence-bound based index policies and finally
 \cite{AuerEtAl02FiniteTime} that popularized a simple sub-optimal index strategy termed \texttt{UCB}
 and most importantly opened the quest for finite-time, as opposed to asymptotic, performance guarantees.  
For the purpose of this paper, we now remind the formal definitions and notations for the stochastic multi-armed bandit problem, following \cite{CaGaMaMuSt2013}.

\paragraph{Quality of a strategy}\label{sub:regret_def}
For each arm $a \in \cA$, let $\mu_a$ be the expectation of the distribution $\nu_a$,
and let $a^\star$ be any optimal arm in the sense that
\[
a^{\star} \in \Argmax_{a \in \cA} \, \mu_a\,.
\]
We write $\mu^\star$ as a short-hand notation for the largest expectation $\mu_{a^{\star}}$
and denote the \emph{gap} of the expected payoff $\mu_a$ of an arm $a$ to $\mu^\star$ as
$\Delta_a = \mu^\star - \mu_a$.
In addition, we denote the number of times each arm $a$ is pulled between the rounds $1$ and $T$
by $N_a(T)$,
\[
N_a(T) \eqdef \sum_{t=1}^T \ind_{ \{ a_t = a \} }\,.
\]
\begin{definition}[Expected regret]
The quality of a strategy is evaluated using the notion of expected regret (or simply, regret) at round $T \geq 1$, defined as
\begin{equation}
\label{eq:defregr}
\kR_T \eqdef \Esp \! \left[ T \mu^{\star} - \sum_{t=1}^T Y_t \right]
= \Esp \! \left[ T \mu^{\star} - \sum_{t=1}^T \mu_{a_t} \right]
= \sum_{a \in\cA} \Delta_a \, \Esp \bigl[ N_a(T) \bigr]\,,
\end{equation}
where we used the tower rule for the first equality. The
expectation is with respect to the random draws of the $Y_t$ according to the $\nu_{a_t}$ and to the possible auxiliary randomization introduced by the decision-making strategy.
\end{definition}

\paragraph{Empirical distributions}\label{sub:empirical}
We denote empirical distributions in two related ways, depending on whether random averages indexed by the global time $t$
or averages of given numbers $t$ of pulls of a given arms are considered. The first series of averages will be
referred to by using a functional notation for the indexation in the global time: $\wh{\nu}_a(t)$,
while the second series will be indexed with the local times $t$ in subscripts: $\wh{\nu}_{a,t}$.
These two related indexations, functional for global times and random averages versus
subscript indexes for local times, will be consistent throughout the paper for all quantities at hand, not only
empirical averages.

\begin{definition}[Empirical distributions]
For each $m \geq 1$, we denote by $\tau_{a,m}$ the round at which arm $a$ was pulled for the $m$--th time, that is
	\[
	\tau_{a,m} = \min \bigl\{ t \in\Nat : \ \ N_a(t) = m \bigr\}\,.
	\]
For each round $t$ such that $N_a(t) \geq 1$, we then define the following two empirical distributions
\[
\wh{\nu}_a(t) = \frac{1}{N_a(t)} \sum_{s=1}^t \delta_{Y_s} \, \ind_{ \{ a_s = a \} }
\quad\text{and}\quad
\wh{\nu}_{a,n} = \frac{1}{n} \sum_{m=1}^n \delta_{X_{a,m}},
\quad \text{where}\quad X_{a,m} \eqdef Y_{\tau_{a,m}}\,.
\]
where $\delta_x$ denotes the Dirac distribution on $x \in \Real$.
\end{definition}

\begin{lemma}\label{lem:empmean}
The random variables $X_{a,m} = Y_{\tau_{a,m}}$, where $m = 1,2,\ldots$, are independent and identically distributed according to $\nu_a$. Moreover, we  have the rewriting
$
\wh{\nu}_{a}(t) = \wh{\nu}_{a,N_a(t)}\,.
$
\end{lemma}
\begin{myproof}{of Lemma~\ref{lem:empmean}}
For means based on local times we consider the filtration $(\cF_t)$, where for all $t \geq 1$, the $\sigma$--algebra $\cF_t$ is generated
by $a_1,Y_1$,\,$\ldots$,\, $a_t,Y_t$. In particular, $a_{t+1}$ and all $N_{a}(t+1)$ are $\cF_t$--measurable. Likewise, $\bigl\{ \tau_{a,m} = t \bigr\}$ is $\cF_{t-1}$--measurable. That is, each
random variable $\tau_{a,m}$ is a (predictable) stopping time. Hence, the result follows by a standard result in probability theory (see, e.g., \citealt[Section~5.3]{chow1988probability}).
\end{myproof}

\section{Boundary crossing probabilities for the generic \KLUCB\ strategy.}\label{sec:generic_klucb}
The first appearance of the \KLUCB\ strategy can be traced at least to \cite{lai1987adaptive} although it was not given an explicit name at that time. It seems the strategy was forgot after the work of \cite{AuerEtAl02FiniteTime} that opened a decade of intensive research on finite-time analysis of bandit strategies and extensions to variants of the problem (\cite{AudibertEtAlUCBV,Audibert2010}, see also \cite{bubeck2012regret} for a survey of relevant variants of bandit problems), until
the work of \cite{HondaTakemura10DMED} shed a novel light on the asymptotically optimal strategies. Thanks to their illuminating work, the first finite-time regret analysis of \KLUCB\ was obtained by \cite{MaillardMunosStoltz11klucb} for discrete distributions,
soon extended to handle exponential families of dimension $1$ as well, in the unifying work of  \cite{CaGaMaMuSt2013}.
However, as we will  see in this paper, we should all be much in dept of the outstanding work of T.L. Lai. regarding the analysis of this index strategy, both asymptotically and in finite-time, as a second look at his papers shows how to bypass the limitations of the state-of-the-art regret bounds for the control of \textit{boundary crossing probabilities} in this context  (see Theorem~\ref{thm:main} below).
Actually, the first focus of the present paper is not stochastic bandits but boundary crossing probabilities, and the bandit setting that we provide here should be considered only as giving a solid motivation for the contribution of this paper.

Let us now introduce formally the \KLUCB\ strategy. We assume that the learner is given a family $\cD\subset\meas$ of probability distributions that satisfies
$\nu_a \in \cD$ for each arm $a\in\cA$, where $\meas$ denotes the set of all probability distributions over $\Real$.
For two distributions $\nu,\nu' \in \meas$, we denote by $\KL(\nu,\nu')$
their Kullback-Leibler divergence and by $E(\nu)$ and $E(\nu')$ their expectations.
(This expectation operator is denoted by $E$ while expectations with respect to underlying
randomizations are referred to as $\Esp$.)

The generic form of the algorithm of interest in this paper is described as Algorithm~\ref{alg:KLUCB}.
It relies on two parameters: an operator $\Pi_\cD$ (in spirit, a projection operator)
that associates with each empirical distribution $\wh{\nu}_a(t)$ an element of the model $\cD$;
and a non-decreasing function $f$, which is typically such that $f(t) \approx \log(t)$.

At each round $t \geq K+1$, a upper confidence bound $U_a(t)$ is associated with the expectation $\mu_a$
of the distribution $\nu_a$ of each arm; an arm $a_{t+1}$ with highest upper confidence bound is then
played.

\begin{algorithm}[h]
\begin{minipage}{\textwidth}
\caption{The \KLUCB\ algorithm (generic form). \label{alg:KLUCB}}
\medskip

\textbf{Parameters:} An operator $\Pi_\cD : \meas \to \cD$; a non-decreasing function $f : \mathbb{N} \to \mathbb{R}$ \smallskip \\
\textbf{Initialization:} Pull each arm of $\{ 1,\ldots, K \}$ once \medskip \\
\textbf{for} each round $t+1$, where $t \geq K$, \textbf{do}{\smallskip

compute for each arm $a$ the quantity
\[
U_a(t) = \sup \biggl\{ E(\nu) : \quad \nu \in \cD \quad \mbox{and} \quad \KL \Bigl( \Pi_\cD \bigl( \wh{\nu}_a(t) \bigr),
\, \nu \Bigr) \leq \frac{f(t)}{N_a(t)} \biggr\}\,;
\]
pick an arm $\quad \displaystyle{a_{t+1} \in \argmax_{a \in \cA} \, U_a(t)}$.
}
\end{minipage}
\end{algorithm}

In the literature, another a variant of \KLUCB\ is introduced where the term $f(t)$ is replaced with $f(t/N_a(t))$. 
We refer to this algorithm as \KLUCBp.
While \KLUCB\ has been analyzed and shown to be provably near-optimal, the variant \KLUCBp\ has not been analyzed yet.

\paragraph{Alternative formulation of \KLUCB}
We wrote the \KLUCB\ algorithm so that the optimization problem resulting from the computation of $U_a(t)$ is easy to handle.
Now, under some assumption, one can rewrite this term, in an equivalent form more suited for the analysis. We refer to \cite{CaGaMaMuSt2013}:

\lemmabox{Rewriting}{rewriting}{
 Under the assumption that
\begin{assumption}\label{ass:implementation}
There is a known interval $\Omega\subset \Real$ with boundary $\mu^- \leq \mu^+$,
for which each  model $\cD_a$ of probability measures is included in $\cP(\Omega)$
and such that $\forall \nu\in\cD_a\forall \mu\in\Omega\setminus\{\mu^+\}$,
\[\inf \, \Bigl\{ \KL(\nu,\nu') : \ \ \nu' \in \cD_a \ \ \mbox{\rm s.t.} \ \ E(\nu') > \mu \Bigr\} = \min \, \Bigl\{ \KL(\nu,\nu') : \ \ \nu' \in \cD_a \ \ \mbox{\rm s.t.} \ \ E(\nu') \geq \mu \Bigr\}\,,\]
\end{assumption}
then the upper bound used by the \KLUCB\ algorithm  satisfies the following equality
\begin{eqnarray*}\label{eq:index}
U_a(t) &=& \max\left\{\mu \in \Omega\setminus\{\mu^+\}: \;  \cK_a\!\Big(\Pi_{a}\left(\hat \nu_a(t)\right), \mu\Big) \leq \frac{f(t)}{N_a(t)}\right\}\,\\
&&\mbox{where} \quad \cK_{a}(\nu_a, \mu^\star) \eqdef \inf_{\nu \in\cD_a: \, E(\nu)> \mu^\star} \KL(\nu_a,\nu)\,.
 \end{eqnarray*}
 Likewise, a similar result holds for\KLUCBp\ but where $f(t)$ is replaced with $f(t/N_a(t))$.
}
\begin{remark}
For instance, this assumption is valid when $\cD_a=\cP([0,1])$ and $\Omega = [0,1]$.
Indeed we can replace the strict inequality with an inequality provided that $\mu<1$ by \cite{HondaTakemura10DMED}, and the infimum
is reached by lower semi-continuity of the $\KL$ divergence and convexity and closure of the set $\{ \nu' \in \cP([0,1]) \ \ \mbox{\rm s.t.} \ \ E(\nu') \geq \mu\}$.
\end{remark}

\paragraph{Using boundary-crossing probabilities for regret analysis}

We continue this warming-up by restating a convenient way to decompose the regret and make appear the \textit{boundary crossing probabilities} that are at the heart of this paper. The following lemma is a direct adaptation from \cite{CaGaMaMuSt2013}:

\lemmabox{From Regret to Boundary Crossing Probabilities}{regretToboundary}{
Let $\epsilon\in\Real^+$ be a small constant such that $\epsilon\in (0,\min\{\,\mu^\star - \mu_a\,,\, a\in\cA\,\})$. For $\mu,\gamma\in\Real$, let us introduce the following set
\begin{eqnarray*}
\cC_{\mu,\gamma} & = & \Bigl\{ \nu'  \in \kM_1(\Real) : \ \ \cK_a( \Pi_a(\nu'),\mu) < \gamma \Bigr\}\,.
\end{eqnarray*}
Then, the number of pulls of a sub-optimal arm $a\in\cA$ by Algorithm \KLUCB\ satisfies
\beqan
\Esp \bigl[ N_T(a) \bigr] \leq 2 + \inf_{n_0\leq T} \bigg\{n_0 + \sum_{n\geq n_0+1}^T \Pr\Bigl\{ \hat \nu_{a,n} \in \cC_{\mu^\star-\epsilon,f(T)/n}\Bigr\} \bigg\}\\
+
\sum_{t=|\cA|}^{T-1}  \underbrace{\Pr \Bigl\{ N_{a^\star}(t) \,\, \cK_{a^\star} \bigl(\Pi_{a^\star}(\hat{\nu}_{a^\star,N_{a^\star}(t)}), \,
\mu^\star -\epsilon\bigr) > f(t) \Bigr\}}_{\text{Boundary Crossing Probability}}\,.
\eeqan
Likewise, the number of pulls of a sub-optimal arm $a\in\cA$ by Algorithm \KLUCBp\ satisfies
\beqan
\Esp \bigl[ N_T(a) \bigr] \leq 2 + \inf_{n_0\leq T} \bigg\{n_0 + \sum_{n\geq n_0+1}^T \Pr\Bigl\{ \hat \nu_{a,n} \in \cC_{\mu^\star-\epsilon,f(T/n)/n}\Bigr\} \bigg\}\\
+
 \sum_{t=|\cA|}^{T-1} \underbrace{\Pr \Bigl\{ N_{a^\star}(t) \,\, \cK_{a^\star} \bigl(\Pi_{a^\star}(\hat{\nu}_{a^\star,N_{a^\star}(t)}), \,
\mu^\star -\epsilon\bigr) > f(t/N_{a^\star}(t) ) \Bigr\}}_{\text{Boundary Crossing Probability}}\,.
\eeqan
}

\begin{myproof}{of Lemma~\ref{lem:regretToboundary}}
The first part of this lemma for \KLUCB\ is proved in \cite{CaGaMaMuSt2013}. The second part that is about \KLUCBp can be proved straightforwardly following the very same lines.
We thus only provide the main steps here  for clarity:
We start by introducing a small $\epsilon>0$ that satisfies $\epsilon < \min\{\,\mu^\star - \mu_a\,,\, a\in\cA\,\}$, and then consider the following inclusion of events:
\[
\bigl\{ a_{t+1} = a \bigr\} \subseteq
\Bigl\{ \mu^\star -\epsilon<U_{a}(t) \text{ and } a_{t+1}=a\Bigr\} \, \cup \,
\Bigl\{ \mu^\star -\epsilon\geq U_{a^\star}(t) \Bigr\}\,;
\]
indeed, on the event $\displaystyle{\bigl\{ a_{t+1} = a \bigr\} \, \cap \,
\Bigl\{ \mu^\star -\epsilon < U_{a^\star}(t) \Bigr\}}\,,
$ we have, $\mu^\star -\epsilon < U_{a^\star}(t) \leq U_{a}(t)$ (where the last
inequality is by definition of the strategy).
Moreover, let us note that
\beqan
\Bigl\{ \mu^\star\! -\!\epsilon<  U_{a}(t) \Bigr\}
\subseteq \Bigl\{\exists \nu' \!\in\! \cD :E(\nu')>\mu^\star\!-\!\epsilon
\text{ and } N_a(t) \,\, \cK_a \bigl(\Pi_a(\hat{\nu}_{a,N_a(t)}), \, \mu^\star \!-\!\epsilon\bigr) \leq f(t/N_a(t)) \Bigr\}\,,\\
 \text{and}\quad\Bigl\{ \mu^\star \!-\!\epsilon \geq U_{a^\star}(t) \Bigr\}
\subseteq \Bigl\{\exists \nu' \!\in\! \cD :  N_{a^\star}(t) \,\, \cK_{a^\star} \bigl( \Pi_{a^\star}(\hat{\nu}_{a^\star,N_{a^\star}(t)}), \,
\mu^\star \!-\!\epsilon\bigr) > f(t/N_{a^\star}(t)) \Bigr\}\,,
\eeqan
since $\cK_a$ is a non-decreasing function in its second argument and $\cK_a\bigl(\nu,E(\nu)\bigr) = 0$
for all distributions $\nu$.
Therefore, this simple remark leads us to the following decomposition
\begin{multline}
\nonumber
\Esp \bigl[ N_T(a) \bigr] \leq 1 + \sum_{t=|\cA|}^{T-1} \Pr \Bigl\{ N_{a^\star}(t) \,\, \cK_{a^\star} \bigl(\Pi_{a^\star}(\hat{\nu}_{a^\star,N_{a^\star}(t)}), \,
\mu^\star -\epsilon\bigr) > f(t/N_{a^\star}(t)) \Bigr\} \\
 + \sum_{t=|\cA|}^{T-1} \Pr \Bigl\{ N_a(t) \,\, \cK_a \bigl( \Pi_a(\hat{\nu}_{a,N_a(t)}), \, \mu^\star -\epsilon\bigr) \leq f(t/N_a(t)) \andAt \Bigr\}\,.
\end{multline}
 The remaining steps of the proof of the result from \cite{CaGaMaMuSt2013}, equation (10) can now be straightforwardly modified to work with $f(t/N_a(t))$ instead of $f(t)$, thus concluding this proof.
\end{myproof}

Lemma~\ref{lem:regretToboundary} shows that two terms need to be controlled in order to derive regret bounds for the considered strategy.
The \textit{boundary crossing probability} term is arguably the most difficult to handle and is the focus of the next sections. The other term involves the probability that an empirical distribution belongs to a convex set, which can be handled  either direclty as in \cite{CaGaMaMuSt2013} or by resorting to finite-time Sanov-type results such as that of \cite[Theorem 2.1 and comments on page 372]{dinwoodie1992mesures}, or its variant from \cite[Lemma~1]{MaillardMunosStoltz11klucb}. For completeness, the exact result from   \cite{dinwoodie1992mesures}
writes
\begin{lemma}[Non-asymptotic Sanov's lemma]
	Let $\cC$ be an open convex subset of $\cP(\cX)$ such that $\quad\Lambda(\cC) = \inf_{\kappa \in \cC} \KL(\kappa,\nu)$ is finite.  Then, for all $t\geq 1$,
	$\qquad \Pr_\nu\{\hat \nu_t \in \cC \} \leq \exp\big(-t\Lambda(\overline{\cC})\big)\qquad$ 
	where $\overline{\cC}$ is the closure of $\cC$.
\end{lemma}

\paragraph{Scope and focus of this work}
We focus on the setting of stochastic multi-armed bandits because this gives a strong and natural motivation for studying boundary crossing probabilities. However, one should understand that the primary goal of this paper is to give credit to the work of T.L. Lai regarding the neat understanding of boundary crossing probabilities
and not necessarily to provide a regret bound for such bandit algorithms as \KLUCB\ or \KLUCBp.
Also, we believe that results on boundary crossing probabilities are useful beyond the bandit problem in hypothesis testing.
Thus, 
and  
in order to avoid obscuring the main result regarding boundary crossing probabilities,
we choose not to provide regret bounds here and to leave them has an exercise for the interested reader;
controlling the remaining term appearing in the decomposition of Lemma~\ref{lem:regretToboundary} is indeed mostly technical and does not seem to require especially illuminating or fancy idea. We refer to \cite{CaGaMaMuSt2013} for an example of bound in the case of exponential families of dimension $1$.

\paragraph{High-level overview of the contribution} We are now ready to explain the main results of this paper.
For the purpose of clarity, we provide them as an informal statement before proceeding with the technical material.

Our contribution is about the behavior the of the \textit{boundary crossing probability} term for exponential families of dimension $K$ when choosing the threshold function $f(x)=\log(x)+\xi\log\log(x)$.
Our result reads as follows.
\textbf{Theorem (Informal statement)}
\emph{Assuming that the observations are generated from a distribution that belongs to an exponential family of dimension $K$ that satisfies some mild conditions, then for 
	any non-negative $\epsilon$ and some class-dependent but fully explicit constants $c,C$  (also depending on $\epsilon$) it holds
\beqan
\Pr \Bigl\{ N_{a^\star}(t) \,\, \cK_{a^\star} \bigl(\Pi_{a^\star}(\hat{\nu}_{a^\star,N_{a^\star}(t)}), \,
	\mu^\star -\epsilon\bigr) > f(t) \Bigr\}&\leq& \frac{C}{t}\log(t)^{K/2-\xi}e^{-c \sqrt{f(t)}}\\
\Pr \Bigl\{ N_{a^\star}(t) \,\, \cK_{a^\star} \bigl(\Pi_{a^\star}(\hat{\nu}_{a^\star,N_{a^\star}(t)}), \,
	\mu^\star -\epsilon\bigr) > f(t/N_{a^\star}(t) ) \Bigr\}&\leq& \frac{C}{t}\log(tc)^{K/2-\xi-1}\,,
\eeqan
where the first inequality holds for all $t$
	and the second one for large enough $t\geq t_c$ where $t_c$ is class dependent  but explicit and "reasonably" small.
}

We provide the rigorous statement in Theorem~\ref{thm:main} and Corollaries~\ref{cor:boundarycrossing}, \ref{cor:boundarycrossingplus} below. The main interest of this result is that it shows how to tune $\xi$ with respect to the dimension $K$ of the family. Indeed, in order to ensure that the probability term is summable in $t$, the bound suggests that $\xi$ should be at least larger than $K/2-1$. 
The case of exponential families of dimension $1$ ($K=1$) is especially interesting, as it supports the fact that both \KLUCB\ and \KLUCBp\ can be tuned using $\xi=0$ (and even negative $\xi$ for \KLUCB). This was observed in numerical experiments in \cite{CaGaMaMuSt2013} although not theoretically supported until now.

The remaining of the paper is organized as follows: Section~\ref{sec:expo} provides the required  background and notations about exponential families,  Section~\ref{sec:boundary}
provides the precise statements as well as previous results, Section~\ref{sec:mainproof}
details the proof of  Theorem~\ref{thm:main}, and finally Section~\ref{sec:tuning}
details the proof of Corollaries~\ref{cor:boundarycrossing} and \ref{cor:boundarycrossingplus}.

\section{General exponential families, properties and examples}\label{sec:expo}
Before focusing on the boundary crossing probabilities, we require a few tools and definitions related to exponential families.
The purpose of this section is thus to present them and prepare for the main result of this paper.
In this section, for a set $\cX\subset\Real$, we consider a multivariate function $F:\cX\to\Real^K$
and denote $\cY = F(\cX)\subset\Real^K$.

\begin{definition}[Exponential families]
 The exponential family generated by the function $F$ and the reference measure $\nu_0$ on the set $\cX$ is
\[
 \cE(F;\nu_0) = \left\{\nu_\theta \in \mathfrak{M}_1(\cX) \,;\,\forall x\in\cX\,\, \nu_\theta(x) = \exp\big(\,\langle \theta, F(x)\rangle - \psi(\theta)\,\big)\nu_0(x) ,\,\, \theta\in \Real^K \right\}\,,
\]
where $\displaystyle{\psi(\theta) \eqdef \log\int_{\cX} \exp\Big(\langle \theta, F(x)\rangle \Big)\nu_0(dx)}$ is the normalization function (aka log-partition function) of the exponential family.
The vector $\theta$ is called the vector of canonical parameters.
The parameter set of the family is the domain $\Theta_\cD \eqdef
\Big\{ \theta \in \Real^K \,;\, \psi(\theta)<\infty \Big\}$, and the invertible parameter set of the family is  $\Theta_I \eqdef \Big\{ \theta \in \Real^K \,;\,  0 < \lambda_{\texttt{MIN}}(\nabla^2\psi(\theta)) \leq \lambda_{\texttt{MAX}}(\nabla^2\psi(\theta)) < \infty\Big\} \subset \Theta_\cD$,
where $\lambda_{\texttt{MIN}}(M)$ and $\lambda_{\texttt{MAX}}(M)$ denote the minimum and maximum eigenvalues of a semi-definite positive matrix $M$.
\end{definition}
\begin{remark}
When $\cX$ is compact, which is the usual assumption in multi-armed bandits ($\cX=[0,1]$) and $F$ is continuous, then we automatically get $\Theta_\cD = \Real^K$.
\end{remark}

In the sequel, we always assume that the family is regular, that is $\Theta_\cD$ has non empty interior.
Another key assumption is that the parameter $\theta^\star$ of the optimal arm belongs to the interior of $\Theta_I$ and is away from its boundary, which essentially avoids degenerate distributions, as we illustrate below.

\smallskip
{\bf Examples }
Bernoulli distributions form an exponential family with $K=1$,
$\cX=\{0,1\}$, $F(x)=x$,$\psi(\theta)=\log(1+e^\theta)$.
The Bernoulli distribution with  mean $\mu$ has parameter $\theta= \log(\mu/(1-\mu))$. Note that $\Theta_\cD=\Real$ and that degenerate distributions with mean $0$ or $1$ correspond to parameters $\pm\infty$. 

Gaussian distributions on $\cX=\Real$ form an exponential family with $K=2$,
$F(x)=(x,x^2)$, and for each $\theta = (\theta_1,\theta_2)$,  $\psi(\theta) = -\frac{\theta_1^2}{4\theta_2} + \frac{1}{2}\log\Big(-\frac{\pi}{\theta_2}\Big)$.
The Gaussian distribution $\cN(\mu,\sigma^2)$
has parameter $\theta = (\frac{\mu}{\sigma^2}, -\frac{1}{2\sigma^2})$. It is immediate to check that $\Theta_\cD = \Real\times\Real_{\star}^{-}$.  Degenerate distributions with variance $0$ correspond to a parameter $\theta$ with both infinite components, while as $\theta$ approaches the boundary $\Real\times\{0\}$, then the variance tends to infinity. It is natural to consider only parameters that correspond to a not too large variance.

\subsection{Bregman divergence induced by the exponential family}\label{sub:Bregman}
An interesting property of exponential families is the following straightforward identity:
\beqan
\forall \theta,\theta'\in\Theta_\cD,\quad \cK(\nu_{\theta},\nu_{\theta'}) = \langle\theta-\theta',\Esp_{X\sim\nu_{\theta}}(F(X))\rangle - \psi(\theta) + \psi(\theta')\,,
\eeqan
In particular, the vector $\Esp_{X\sim\nu_{\theta}}(F(X))$ is called the vector of \textit{dual (or expectation) parameters}. It is equal to the vector
$\nabla \psi(\theta)$. Now, we write $\cK(\nu_{\theta},\nu_{\theta'}) = \cB^\psi(\theta,\theta')$, where we introduced the Bregman divergence
with potential function $\psi$ defined by
\beqan
\cB^\psi(\theta,\theta')  \eqdef  \psi(\theta') - \psi(\theta) -\langle\theta'-\theta,\nabla \psi(\theta)\rangle  \,.
\eeqan

Thus, if $\Pi_{a}$ is chosen to be the projection on the exponential family  $\cE(F;\nu_0)$,
and $\nu$ is a distribution with projection given by $\nu_\theta = \Pi_{a}(\nu)$, then we can rewrite the definition of $\cK_a$ in the simpler form
\beqa
\label{eqn:bregmanKinf}
\cK_{a} \bigl(\Pi_{a}(\nu), \, \mu\bigr)  = \inf\big\{\,\cB^\psi(\theta,\theta') \,;\, \Esp_{\nu_{\theta'}}(X) > \mu \,\big\}\,.
\eeqa

We continue by providing a powerful rewriting of the Bregman divergence. 

\lemmabox{Bregman duality}{bregdual}{
Let $\Phi(\eta) = \psi(\theta^\star+\eta)-\psi(\theta^\star)$, and its Fenchel-Legendre dual given by
\beqan
\Phi^\star(y) = \sup_{\eta\in\Real^K}
\langle \eta, y\rangle  - \psi(\theta^\star+\eta)+\psi(\theta^\star)\,.
\eeqan
Then, for all $\theta^\star\in\Theta_\cD$ and $\eta\in\Real^K$ such that
$\theta^\star+\eta\in\Theta_\cD$, it holds
\beqan
\log \Esp_{\theta^\star} \exp\bigg( \langle \eta, F(X)\rangle \bigg) = \Phi(\eta)\,.
\eeqan
Further, for all $F\in \nabla \psi(\Theta_\cD)$ with $F= \nabla \psi(\theta)$ for some $\theta\in\Theta_\cD$, then $\quad\Phi^\star(F) = \cB^\psi(\theta,\theta^\star)$.
}

\lemmabox{Bregman and Smoothness}{taylor2}{
We have on the one hand
\beqan
  \cB^{\psi}(\theta,\theta')  \leq \frac{|\theta-\theta'|^2}{2} \sup \{\lambda_{\texttt{MAX}}(\nabla^2 \psi(\tilde \theta))\,;\,\tilde \theta \in [\theta,\theta'] \}\,,\\
  |\nabla \psi(\theta) - \nabla\psi(\theta')| \leq \sup \{\lambda_{\texttt{MAX}}(\nabla^2 \psi(\tilde \theta))\,;\,\tilde \theta \in [\theta,\theta'] \} | \theta-\theta'|\,,
\eeqan
and on the other hand 
\beqan
  \cB^{\psi}(\theta,\theta')  \geq \frac{|\theta-\theta'|^2}{2} \inf \{\lambda_{\texttt{MIN}}(\nabla^2 \psi(\tilde \theta))\,;\,\tilde \theta \in [\theta,\theta'] \}\,,\\
  |\nabla \psi(\theta) - \nabla\psi(\theta')| \geq \inf \{\lambda_{\texttt{MIN}}(\nabla^2 \psi(\tilde \theta))\,;\,\tilde \theta \in [\theta,\theta'] \} | \theta-\theta'|\,,
\eeqan
where $\lambda_{\texttt{MAX}}(\nabla^2 \psi(\tilde \theta))$ and $\lambda_{\texttt{MIN}}(\nabla^2 \psi(\tilde \theta))$ are the largest and smallest eigenvalue of
$\nabla^2 \psi(\tilde \theta)$.
}

\begin{myproof}{of Lemma~\ref{lem:bregdual}}
	The second equality holds by simple algebra. Now
	the first equality is immediate, since
	
	\vspace{-6mm}
	\beqan
	\log \Esp_{\theta^\star} \exp(\langle \eta, F(X)\rangle )&=&
	\log \int\exp(\langle \eta, F(x)\rangle + \langle{\theta^\star}, F(x)\rangle - \psi(\theta^\star))\nu_0(dy)\\
	&=& \psi(\eta+\theta^\star) - \psi(\theta^\star)\,.
	\eeqan
	
	\vspace{-7mm}
	
\end{myproof}
\begin{myproof}{of Lemma~\ref{lem:taylor2}}
We have by definition that $\quad
\cB^{\psi}(\theta,\theta') = \psi(\theta) - \psi(\theta') - \langle\theta - \theta', \nabla \psi (\theta')\rangle\,.\quad$\\
Then, by a Taylor expansion, there exists $\tilde \theta' \in [\theta,\theta']$ such that
\beqan
\psi(\theta)  = \psi(\theta')  + \langle\theta - \theta', \nabla \psi (\theta')\rangle + \frac{1}{2}(\theta - \theta')^T\nabla^2 \psi (\tilde \theta)(\theta - \theta')\,.
\eeqan
Likewise, there exists $\tilde \theta \in [\theta,\theta']$ such that
$\quad
\nabla\psi(\theta)  = \nabla\psi(\theta')  + \nabla^2 \psi (\tilde \theta)(\theta - \theta')\,.$
\end{myproof}

\subsection{Dual formulation of the optimization problem}
Using Bregman divergences enables to rewrite the $K$-dimensional optimization problem \eqref{eqn:bregmanKinf} in a slightly more convenient form thanks to a dual formulation. Indeed introducing a Lagrangian parameter $\lambda \in \Real^+$ and using Karush-Kuhn-Tucker conditions, one gets the following necessary optimality conditions
\beqan
\nabla \psi(\theta') - \nabla \psi(\theta) - \lambda \partial_{\theta'} \Esp_{\nu_{\theta'}}(X) = 0, \text{ with}\\
\lambda ( \mu - \Esp_{\nu_{\theta'}}(X)) = 0, \quad \lambda \geq 0, \quad \Esp_{\nu_{\theta'}}(X) \geq \mu\,,
\eeqan
and by definition of the exponential family, we can make use of the fact that
\beqan
\partial_{\theta'} \Esp_{\nu_{\theta'}}(X) = \Esp_{\nu_{\theta'}}(XF(X)) - \Esp_{\nu_{\theta'}}(X)\nabla\psi(\theta') \in\Real^K\,,
\eeqan
where we remember that $X\in\Real$ and $F(X) \in \Real^K$. Combining these two equations, we obtain the system
\beqa\label{eqn:KKTgeneric}
\begin{cases}
&\nabla \psi(\theta') (1+\lambda \Esp_{\nu_{\theta'}}(X) ) - \nabla \psi(\theta) - \lambda \Esp_{\nu_{\theta'}}(XF(X)) = 0 \in\Real^K,\\
& \text{with }\lambda ( \mu - \Esp_{\nu_{\theta'}}(X)) = 0, \quad \lambda \geq 0, \quad \Esp_{\nu_{\theta'}}(X) \geq \mu\,.
\end{cases}
\eeqa
For minimal exponential family, this system admits for each fixed $\theta,\mu$ a unique solution in $\theta'$, that we write for clarity $\theta(\lambda^\star;\theta,\mu)$ to indicate
its dependency with the optimal value $\lambda^\star$ of the dual parameter as well as the constraints.

\begin{remark} 
	For $\theta\in\Theta_I$, when the optimal value of $\lambda$ is $\lambda^\star=0$, then
it means that  $\nabla \psi(\theta') = \nabla \psi(\theta)$ and thus $\theta'=\theta$, which is only possible if $\Esp_{\nu_\theta}(X) \geq \mu$. Thus whenever $\mu>\Esp_{\nu_\theta}(X)$, the dual constraint is active, i.e. $\lambda>0$, and we get the vector equation
\beqan\label{eqn:KKT}
\nabla \psi(\theta')(1+\lambda \mu) - \nabla \psi(\theta) - \lambda\Esp_{\nu_{\theta'}}(XF(X)) = 0\,\,\text{ and }\,\, \Esp_{\nu_{\theta'}}(X) =  \mu\,.
\eeqan
\end{remark}

{\bf The example of discrete distributions}
In many cases, the previous optimization problem reduces to a simpler one-dimensional optimization problem, where we optimize over the dual parameter $\lambda$.  We illustrate this phenomenon on a family of discrete distributions. Let $\bX=\{x_1,\dots,x_K,x_\star\}$ be a set of distinct real-values. Without loss of generality, assume that $x_\star > \max_{k\leq K} x_k$.
The family of distributions $p$ with support in $\bX$ is a specific $K$-dimensional family. Indeed, let $F$ be the feature function with $k^{th}$ component $F_k(x) = \indic{x=x_k}$, for all $k\in\{1,\dots, K\}$.
Then the parameter $\theta = (\theta_k)_{1\leq k\leq K}$ of the distribution $p=p_\theta$ has components $\theta_k = \log(\frac{p(x_k)}{p(x_\star)})$ for all $k\neq 0$. Note that $p(x_k)= \exp(\theta_k - \psi(\theta))$ for all $k\neq 0$,
and $p(x_0) = \exp(-\psi(\theta))$. It then comes  $\psi(\theta) = \log(\sum_{k=1}^Ke^{\theta_k} +1)$, $\nabla \psi(\theta) = (p(x_1),\dots,p(x_K))^\top$ and 
 $\Esp(XF_k(X)) = x_k p_\theta(x_k)$. Further, 
 $\Theta_\cD = (\Real\cup\{-\infty\})^K$ and $\theta\in\Theta_\cD$
 corresponds to the condition $p_\theta(x_\star)>0$.
Now, for a non trivial value $\mu$ such that $\Esp_{p_\theta}(X)<\mu<x_\star$,  it can be readily checked that the system \eqref{eqn:KKTgeneric} specialized to this family is equivalent (with no surprise) to the one considered for instance in \cite{HondaTakemura10DMED} for discrete distributions. 
After some tedious but simple steps detailed in \cite{HondaTakemura10DMED}, one obtains the following easy-to-solve  \textit{one-dimensional} optimization problem  (see also \cite{CaGaMaMuSt2013}), although the family is of dimension $K$:
\beqan
\cK_{a} \bigl(\Pi_{a}(\nu), \, \mu\bigr) =
\cK_{a} \bigl(\nu_\theta, \, \mu\bigr) =  \sup \Big\{  \sum_{x\in\bX} p_{\theta}(x)\log\Big(1-\lambda \frac{x-\mu}{x_\star-\mu}\Big) \,;\, \lambda \in [0,1]\Big\}\,.
\eeqan

\subsection{Empirical parameter and definition}
In this section we discuss the well-definition of the empirical parameter corresponding to the projection of the empirical distribution on the exponential family. While this is innocuous for most settings, in full generality, one needs
to take some specific care to ensure that all the objects we deal with are well-defined and that all parameters $\theta$ we talk about indeed belong to the set
$\Theta_D$ (or better $\Theta_I$).

An important property is that if the family is regular, then $\nabla\psi(\Theta_\cD)$ is an open set that coincides with the interior of realizable values of $F(x)$ for $x\sim\nu$ for any $\nu$ absolutely continuous with respect to $\nu_0$.
In particular, by convexity of the set
$\nabla\psi(\Theta_\cD)$ this means that the empirical average 
$\frac{1}{n}\sum_{i=1}^nF(X_i)\in\Real^K$ belongs to $\overline{\nabla\psi(\Theta_\cD)}$ for all $\{X_i\}_{i\leq n}\sim\nu_\theta$ with $\theta\in\Theta_\cD$.
Thus, for the observed samples
$X_1,\dots,X_n \in \cX$ coming from $\nu_{a^\star}$,
the projection $\Pi_{a^\star}(\hat{\nu}_{a^\star,n})$ on the family can be represented by a sequence $\{\hat \theta_{n,m}\}_{m\in\Nat} \in \Theta_\cD$ such that
\beqan
\nabla \psi(\hat \theta_{n,m}) \stackrel{m}{\to}\hat F_n \,\text{ where }\, \hat F_n \eqdef \frac{1}{n}\sum_{i=1}^n F(X_i) \in \Real^K\,.
\eeqan

In the sequel, we want to ensure that
provided that $\nu_{a^\star} = \nu_{\theta^\star}$ with $\theta^\star\in\mathring{\Theta}_I$, then we also have
$\hat F_n \in \nabla\psi(\mathring{\Theta}_I)$, which means that there is a unique $\hat \theta_n \in \mathring{\Theta}_I$ such that $\nabla \psi(\hat \theta_{n}) = \hat F_n$, or equivalently $\hat \theta_n = \nabla \psi^{-1} (\hat F_n)$. 
To this end, we assume that $\theta^\star$ is away from the boundary of $\Theta_I$. In many cases, it is then sufficient to assume that 
$n$ is larger than a small constant (roughly $K$) to ensure that we can find a unique $\hat \theta_n \in \mathring{\Theta}_I$ such that $\nabla \psi(\hat \theta_{n}) = \hat F_n$.

{\bf Example}
Let us consider Gaussian distributions on $\cX=\Real$, with $K=2$.
We consider a parameter $\theta^\star= (\frac{\mu}{\sigma^2},-\frac{1}{2\sigma^2})$ corresponding to a Gaussian finite mean $\mu$ and positive variance $\sigma^2$. Now, for any $n\geq 2$, the empirical mean $\hat \mu_n$ is finite and
the empirical variance $\hat \sigma^2_n$ is positive, and thus $\theta_n
= \nabla \psi^{-1}(\hat F_n)$ is well-defined. 

The case of Bernoulli distributions is interesting as it shows a slightly different situation.
Let us consider a parameter $\theta^\star = \log(\mu/(1-\mu))$ corresponding to a Bernoulli distribution with mean $\mu$. 
Before $\hat F_n$ can be mapped to a point in $\mathring{\Theta}_I=\Real$, one needs to wait that the number of observations for both $0$ and $1$ is positive. Whenever $\mu\in(0,1)$,  the probability that this does not happen is controlled by
$\Pr( n_0(n) =0 \text{ or } n_1(n) = 0) = \mu^n + (1-\mu)^n \leq 2 \max(\mu,1-\mu)^n,$ where $n_x(n)$ denotes the number of observations of symbol $x\in\{0,1\}$
after $n$ samples. For $\mu\geq 1/2$, the later quantity is less than $\delta_0\in(0,1)$ for $n \geq \frac{\log(2/\delta_0)}{\log(1/\mu)}$, which depends on the probability level $\delta_0$ and cannot be considered to be especially small when $\mu$ is close  \footnote{This also suggests to replace $\hat F_n$ with a Laplace or a Krichevsky-Trofimov estimate that provide initial bonus to each symbol and, as a result, maps any $\hat F_n$, for $n \geq 0$ to a parameter in $\hat \theta_n \in\Real$.} to $1$.
That said, even when the parameter $\hat \theta_n$ does not belong to $\Real$, the event $n_0(n)=0$ corresponds to having empirical mean equal to $1$. This is a favorable situation since any optimistic algorithm should pull the corresponding arm.
Thus, we one only need to control $\Pr(n_1(n) = 0) = (1-\mu)^n$, which is less than $\delta_0\in(0,1)$ for $n \geq \frac{\log(1/\delta_0)}{\log(1/(1-\mu))}$, which is essentially a constant.
As a matter of illustration, when $\delta=10^{-3}$ and $\mu=0.9$, this condition is met for $n\geq 3$.

\medskip
Following the previous discussion, in the sequel we consider that $n$ is always large enough so that   $\hat \theta_n = \nabla \psi^{-1} (\hat F_n) \in  \mathring{\Theta}_I$ can be uniquely defined. We now discuss the separation between the parameter and the boundary more formally, and for that purpose introduce the following definition.

\begin{definition}[Enlarged parameter set]
Let $\Theta\subset \Theta_\cD$ and some constant $\rho>0$. The enlargement of size $\rho$ of $\Theta$ in Euclidean norm (aka $\rho$-neighborhood) is defined by \beqan
\Theta_\rho \eqdef \Big\{ \theta \in \Real^K \,;  \inf_{\theta'\in \Theta_\cD} |\theta-\theta'| < \rho\,\Big\}\,.
\eeqan
For each $\rho$ such that $\Theta_\rho\subset\Theta_I$, we further introduce  the quantities 
\beqan
v_\rho = v_{\Theta_\rho} \eqdef \inf_{\theta \in \Theta_\rho} \lambda_{\texttt{MIN}}(\nabla^2 \psi(\theta)),&&
V_\rho= V_{\Theta_\rho} \eqdef \sup_{\theta \in \Theta_\rho} \lambda_{\texttt{MAX}}(\nabla^2 \psi(\theta))\,.
\eeqan
\end{definition}

Using the notion of enlarged parameter set, we highlight an especially useful property to prove concentration inequalities, summarized in the following result

\lemmabox{Log-Laplace control}{logLaplacecontrol}{
Let $\Theta\subset\Theta_\cD$ be a convex set and $\rho>0$ such that
$\theta^\star\in\Theta_{\rho} \subset\Theta_I$. Then, for all 
$\eta\in\Real^K$ such that $\theta^\star+\eta\in \Theta_{\rho}$, it holds
\beqan
\log \Esp_{\theta^\star} \exp(\eta^\top F(X) )&\leq& \eta^\top \nabla \psi(\theta^\star) +
\frac{V_{\rho} }{2}\norm{\eta}^2\,.
\eeqan
}

\begin{myproof}{of Lemma~\ref{lem:logLaplacecontrol}}
Indeed, it holds by simple algebra
\beqan
\log \Esp_{\theta^\star} \exp(\eta^\top F(X) )&=& \psi(\theta^\star+\eta) - \psi(\theta^\star)\\
&\leq& \eta^\top \nabla \psi(\theta^\star) +
\max_{\theta \in H(\theta^\star+\eta,\theta^\star)}
\frac{1}{2}\eta^\top \nabla^2\psi(\theta)\eta\\
&\leq& \eta^\top \nabla\psi(\theta^\star) +
\frac{V_{\rho} }{2}\norm{\eta}^2\,,
\eeqan
where $H(\theta,\theta') = \{ \alpha \theta + (1-\alpha)\theta', \alpha\in[0,1]\}$.
The equality holds by definition and basic rewriting.
In the inequalities, we used that $\Theta_\rho$ is convex as an enlargement of a convex set, and thus that $H(\eta+\theta^\star,\theta^\star)\subset \Theta_\rho$.
\end{myproof}

In the sequel, we are interested in sets $\Theta$
such that $\Theta_\rho \subset \mathring{\Theta}_I$ for some specific $\rho$. This comes essentially from the fact that we require some room around $\Theta$ and $\Theta_I$ to ensure all quantities remain finite and well-defined.
Before proceeding, it is convenient to introduce the notation
$d(\Theta',\Theta) = \inf_{\theta\in\Theta,\theta'\in\Theta'} \norm{\theta-\theta'}$, as well as the Euclidean ball 
$B(y,\delta) = \{y'\in\Real^K: ||y'-y||\leq \delta \}$. Using these notations, 
the following lemma whose proof is immediate provides conditions for which
all future technical considerations are satisfied.

\lemmabox{Well-defined parameters}{Welldefined}{
Let $\theta^\star \in \mathring{\Theta}_I$ and $\rho^\star = d(\{\theta^\star\}, \Real^K\setminus\Theta_I)>0$.  Now for any convex set $\Theta\subset \Theta_I$ such that $\theta^\star\in\Theta$ and $d(\Theta, \Real^K\setminus\Theta_I)= \rho^\star$, and any $\rho<\rho^\star/2$, it holds $\Theta_{2\rho} \subset \mathring{\Theta}_I$.

Further, for any $\delta$ such that $\hat F_n\! \in\! B( \nabla\psi(\theta^\star), \delta)\! \subset\! \nabla \psi(\Theta_{\rho})$,   $\exists\hat \theta_n\! \in\! \Theta_{\rho}\! \subset\! \mathring{\Theta}_I$ such that  $\nabla \psi(\hat \theta_n)\! =\! \hat F_n$.}

In the sequel, we will restrict our analysis to the slightly more restrictive case when $\hat \theta_n \in \Theta_{\rho}$ with $\Theta_{2\rho}\subset \mathring{\Theta}_I$. This is mostly for convenience and avoid dealing with rather specific situations.

\begin{remark}
Again let us remind that when $\cX$ is compact and $F$ is continuous,
then $\Theta_I=\Theta_\cD=\Real^K$. 
\end{remark}

\paragraph{Illustration}
We now illustrate the definition of $v_\rho$ and $V_\rho$.
For Bernoulli distributions with parameter $\mu\in[0,1]$,  $\nabla \psi(\theta) = 1/(1+e^{-\theta})$ and $\nabla^2\psi(\theta) = e^{-\theta}/(1+e^{-\theta})^2 = \mu(1-\mu)$. Thus, $v_\rho$ is away from $0$ whenever $\Theta_\rho$ excludes the means $\mu$ close to $0$ or $1$, and $V_\rho\leq 1/4$.

Now for a family of Gaussian distributions with unknown mean and variance,
$\psi(\theta)= - \frac{\theta_1^2}{4\theta_2} + \frac{1}{2}\log\big(\frac{-\pi}{\theta_2}\big)$, where $\theta = (\frac{\mu}{\sigma^2},-\frac{1}{2\sigma^2})$.
Thus, $\nabla\psi(\theta) = (-\frac{\theta_1}{2\theta_2}, \frac{\theta_1^2}{4\theta_2^2} -  \frac{1}{2\theta_2} )$, and 
$\nabla^2\psi(\theta) = (-\frac{1}{2\theta_2}, \frac{\theta_1}{2\theta_2^2} ; \frac{\theta_1}{2\theta_2^2},  -\frac{\theta_1^2}{2\theta_2^3} + \frac{1}{2\theta_2^2})
= 2\mu\sigma^2( \frac{1}{2\mu},1;1,2\mu +\frac{\sigma^2}{\mu})$. The smallest eigenvalue is larger than $\sigma^4/(1/2+\sigma^2 + 2\mu^2)$ and 
the largest is upper bounded by
$\sigma^2(1+2\sigma^2+4\mu^2)$, which enables to control $V_\rho$ and $v_\rho$.

\section{Boundary crossing for $K$-dimensional exponential families}\label{sec:boundary}
In this section, we now study the boundary
crossing probability term appearing in Lemma~\ref{lem:regretToboundary}
for a $K$-dimensional exponential family $\cE(F;\nu_0)$.
We first provide an overview of the existing results before detailing our main contribution. As explained in the introduction, the key technical tools that enable to obtain the novel results were already known three decades ago, and thus even though the novel result is impressive due to its generality and tightness, it should be regarded as a modernized version of an existing, but almost forgotten result, that enables to solve a few long-lasting open questions as a by-product.

\subsection{Previous work on boundary-crossing probabilities}

The existing results used in the bandit literature about boundary-crossing probabilities are restricted to a few specific cases. For instance in \cite{CaGaMaMuSt2013}, the authors provide the following control

\theorembox{\KLUCB}{mainKL-weak}{
In the case of canonical (that is $F(x)=x$) exponential families of dimension $K=1$, then 
for a function $f$ such that $f(x) = \log(x) + \xi \log\log(x)$, then it holds for all $t>A$
\beqan
\Pr_{\theta^\star} \Bigl\{ \bigcup_{n=1}^{t-A+1}  n \,\,
\cK_{a^\star} \bigl( \Pi_{a^\star}(\hat{\nu}_{a^\star,n}), \, \mu^\star\bigr) > f\big(t\big) \cap \mu_{a^ \star} > \hat \mu_{a^\star,n}\Bigr\} \leq e\lceil f(t)\log(t)\rceil e^{-f(t)}\,.
\eeqan
Further, in the special case of distributions with finitely many $K$ atoms, it holds for all $t>A, \epsilon>0$
\beqan
\Pr_{\theta^\star} \Bigl\{ \bigcup_{n=1}^{t-A+1}  n
\cK_{a^\star} \bigl( \Pi_{a^\star}(\hat{\nu}_{a^\star,n}), \, \mu^\star -\epsilon \bigr) > f\big(t\big) \Bigr\} \leq e^{-f(t)}\Big(3e+2+4\epsilon^{-2} + 8e\epsilon^{-4}\Big)\,.
\eeqan
}\\
In contrast in \cite{lai1988boundary}, the authors provide an asymptotic control in
the more general case of exponential families of dimension $K$ with some basic regularity condition, as we explained earlier. 
We now restate this beautiful  result from \cite{lai1988boundary} in a way that is suitable for a more direct comparison with other results. The following holds:

\theorembox{Lai, 88}{mainLai}{
Let us consider an exponential family of dimension $K$.
Define for $\gamma>0$ the cone $\cC_\gamma(\theta) = \{ \theta'\in\Real^K : \langle \theta', \theta \rangle \geq \gamma |\theta||\theta'|\}$.
Then, for a function $f$ such that $f(x) = \alpha\log(x) + \xi\log\log(x)$ it holds for all $\theta^\dagger\in\Theta$ such that $|\theta^\dagger-\theta^\star|^2 \geq \delta_t$, where $\delta_t\to0$, $t\delta_t\to\infty$ as $t\to\infty$,
\beqan
\lefteqn{
\hspace{-3cm}
\Pr_{\theta^\star} \Bigl\{\bigcup_{n=1}^t \hat \theta_n \in \Theta_\rho \,\cap\,
n\cB^\psi(\hat \theta_n,\theta^\dagger) \geq f\Big(\frac{t}{n}\Big)
\,\cap\, \nabla \psi(\hat \theta_n) - \nabla \psi(\theta^\dagger)
\in \cC_\gamma(\theta^\dagger-\theta^\star) \Bigr\}}\\
&\stackrel{t\to\infty}{=}& O\bigg(t^{-\alpha}|\theta^\dagger-\theta^\star|^{-2\alpha} \log^{-\xi-\alpha + K/2}(t|\theta^\dagger-\theta^\star|^2)\bigg)\\
&=& O\bigg(e^{-f(t|\theta^\dagger-\theta^\star|^2)}\log^{-\alpha + K/2}(t|\theta^\dagger-\theta^\star|^2)\bigg)\,.
\eeqan
}

\medskip
{\bf Discussion}
The quantity $\cB^\psi(\hat \theta_n,\theta^\dagger)$
is the direct analog  of $\cK_{a^\star} \bigl( \Pi_{a^\star}(\hat{\nu}_{a^\star,n}),\mu^\star-\epsilon)$
in Theorem~\ref{thm:mainKL-weak}. Note however that $f(t/n)$ replaces the larger quantity $f(t)$, which means that Theorem~\ref{thm:mainLai} controls a larger quantity than Theorem~\ref{thm:mainKL-weak}, and is thus in this sense stronger.
It also holds for general exponential families of dimension $K$.
Another important difference is the order of magnitude of the right hand side terms of both theorems.
Indeed, since $e\lceil f(t)\log(t)\rceil e^{-f(t)} =O(\frac{\log^{2-\xi}(t)+\xi\log(t)^{1-\xi}\log\log(t)}{t})$, Theorem~\ref{thm:mainKL-weak} requires that $\xi>2$ in order that this term is $o(1/t)$, and $\xi>0$ for the second term of Theorem~\ref{thm:mainKL-weak}.
In contrast, Theorem~\ref{thm:mainLai}
shows that it is enough to consider $f(x) = \log(x) + \xi\log\log(x)$ with $\xi>K/2-1$ to ensure a $o(1/t)$ bound. For $K=1$, this means we can even use $\xi>-1/2$ and in particular $\xi=0$, which corresponds to the value they recommend in the experiments.

Thus, Theorem~\ref{thm:mainLai} improves in three ways over Theorem~\ref{thm:mainKL-weak}: it is an extension to dimension $K$, it provides a bound for $f(t/n)$
(and thus for \KLUCBp) and not only $f(t)$, and finally allows for smaller values of $\xi$.
These improvements are partly due to the fact 
Theorem~\ref{thm:mainKL-weak} controls a concentration with respect to $\theta^\dagger$, not $\theta^\star$,
which takes advantage of the fact there is some gap when going from $\mu^\star$ to distributions with mean $\mu^\star-\epsilon$. The proof of Theorem~\ref{thm:mainLai} directly takes advantage of this, contrary to that of the first part of Theorem~\ref{thm:mainKL-weak}.

On the other hand, Theorem~\ref{thm:mainLai} is only asymptotic whereas  Theorem~\ref{thm:mainKL-weak} holds for finite $t$. Furthermore, we notice two restrictions on the control event. First, it requires $\hat \theta_n \in \Theta_\rho$, but we showed  in the previous section that this is a minor restriction.
Second, there is the restriction to a cone $\cC_\gamma(\theta^\dagger-\theta^\star)$ which simplifies the analysis, but is a more dramatic restriction. This restriction cannot be removed trivially as it can be seen from the complete statement of \cite[Theorem 2]{lai1988boundary} that the right hand-side blows up to $\infty$ when $\gamma\to0$. As we will see, it is possible to overcome this restriction by resorting to a smart covering of the space with cones, and sum the resulting terms via a union bound over the covering.
We explain the precise way of proceeding 
in the proof of Theorem~\ref{thm:main} in section~\ref{sec:mainproof}.

\medskip
{\bf Hint at proving the first part of Theorem~\ref{thm:mainKL-weak}} 
We believe it is interesting to give some hint the proof of the first part of Theorem~\ref{thm:mainKL-weak}, as it involves an elegant step, despite relying quite heavily on two specific properties of the canonical exponential family of dimension $1$.
Indeed in the special case of the canonical one-dimensional family (that is $K=1$ and  $F_1(x)=x\in\Real$), $\hat F_n = \frac{1}{n}\sum_{i=1}^n X_i$ coincides with the empirical mean and 
it can be shown that  $\Phi^\star(F)$ is strictly decreasing
on $(-\infty,\mu^\star]$. Thus for any $F\leq \mu^\star$, it holds
\beqa
\bigg\{ \hat F_n \leq \mu^\star \cap \Phi^\star(\hat F_n) \geq \Phi^\star(F) \bigg\} \subset
\bigg\{ \hat F_n \leq F \bigg\}\,.\label{eqn:PhistarDec}
\eeqa
Further, using the notations of Section~\ref{sub:Bregman}, it also holds in that case
$\cK_{a^\star} \bigl( \Pi_{a^\star}(\hat{\nu}_{a^\star,n}), \, \mu^\star\bigr) 
 = \cB^\psi(\hat \theta_n, \theta^\star) = \Phi^\star(\hat F_n)$, where $\hat \theta_n = \dot \psi^{-1}(\hat F_n)$ is uniquely defined. A second non-trivial property that is shown in \cite{CaGaMaMuSt2013} is that for all $F\leq \mu^\star$, we can localize the supremum as 
\beqa
\Phi^\star(F) = \sup\bigg\{xF  - \Phi(x):
x <0 \text{ and } x F -\Phi(x)>0
\bigg\}\,.\label{eqn:Supremum}
\eeqa

Armed with these two properties, the proof reduces almost trivially to the following elegant lemma:

\lemmabox{Dimension 1}{dim1KL}{
Consider a canonical one-dimensional family (that is $K=1$ and  $F_1(x)=x\in\Real$). Then, for all $f$ such that $f(t/n)/n$ is non-increasing in $n$,
\beqan
\Pr_{\theta^\star}\Big\{\bigcup_{m\leq n < M} \,\, \cB^\psi(\hat \theta_n,\theta^\star) \geq f(t/n)/n \Big\}&\leq&\exp\bigg(-\frac{m}{M}f(t/M)\bigg)\,.
\eeqan
}

 This lemma, whose proof is provided in the appendix for the interested reader and is directly adapted from the proof of 
Theorem~\ref{thm:mainKL-weak}. The first statement of
Theorem~\ref{thm:mainKL-weak} is obtained by a peeling argument, 
using $m/M = (f(t)-1)/f(t)$. However this argument does not seem to extend  nicely to using $f(t/n)$, which explains why there is no statement regarding this threshold.

\subsection{Main results and contributions}
In this section, we now provide several results on boundary crossing probabilities, that we prove in details in the next section.
We first provide a non-asymptotic bound
with explicit terms for the control of the boundary crossing probability term.
We then provide two corollaries that can be used directly for the analysis of \KLUCB\ and \KLUCBp and that better highlight the asymptotic scaling of the bound with $t$, which helps seeing the effect of the parameter $\xi$ on the bound.

 \theorembox{Boundary crossing for exponential families}{main}{
Let $\epsilon<\min_{a\in\cA : \mu_a<\mu^\star}(\mu^\star-\mu_a)$,
and define $\rho_\epsilon = \inf\{ ||\theta'-\theta|| : \mu_{\theta'}= \mu^\star-\epsilon, \mu_\theta=\mu^\star  \}$. Let $\rho^\star = d(\{\theta^\star\},\Real^K\setminus \Theta_I)$
 	and $\Theta\subset \Theta_\cD$ be a set  such that  $\theta^\star \in\Theta$ and $d(\Theta,\Real^K\setminus \Theta_I) = \rho^\star$. Thus $\theta^\star\in\Theta\subset\Theta_{\rho} \subset \mathring{\Theta}_I$ for each $\rho<\rho^\star$.
 	Assume that $n\to f(t/n)/n$ is non-increasing and 
 	$n\to nf(t/n)$ is non-decreasing. 	
  	Then, for every $b>1, p,q,\eta \in [0,1]$,
  	and $n_i=b^i$ if $i<I_t = \lceil \log_b(qt)\rceil$, $n_{I_t}=t+1$, it holds
 	\beqan
 	\lefteqn{
 		\Pr_{\theta^\star}\Big\{\bigcup_{1\leq n \leq t}\hat \theta_n\in\Theta_\rho\cap \cK_{a^\star}(\Pi_{a^\star}(\hat \nu_{a^\star,n}),\mu^\star-\epsilon) \geq f(t/n)/n\Big\}}\\
 	&\leq& C(K,b,\rho,p,\eta)
 	\sum_{i=0}^{I_t-1}
 	\exp\bigg(
 	- n_i\rho_\epsilon^2\alpha^2
 	- \rho_\epsilon\chi\sqrt{n_if(t/n_i)}- f\Big(\frac{t}{n_{i+1}\!-\!1}\Big)\bigg)f\Big(\frac{t}{n_{i+1}\!-\!1}\Big)^{K/2}\,,
 	\eeqan
 	where we introduced the constants 	$\alpha =\eta \sqrt{v_\rho/2}$, $\chi = p\eta \sqrt{2v_\rho^2/V_\rho}$ and 
 	\beqan
 	C(K,b,\rho,p,\eta) =  C_{p,\eta,K}\Big(2\frac{\omega_{p,K\!-\!2}}{\omega_{\max\{p,\frac{2}{\sqrt{5}} \},K\!-\!2}}\max\Big\{\frac{2bV_\rho^4}{p\rho^2 v^6_\rho} , \frac{V_\rho^3}{v_\rho^4},\frac{b^2V_\rho^5}{pv_\rho^6(\frac{1}{2}\!+\!\frac{1}{K})} \Big\}^{K/2}
 	+1\Big)\,,
 	\eeqan
 	where  $C_{p,\eta,K}$ is the cone-covering number of $\nabla\psi\big(\Theta_\rho\setminus\cB_2(\theta^\star,\rho_\epsilon)\big)$ with minimal angular separation $p$ and not intersecting the set $\nabla\psi\big(\Theta_\rho\setminus\cB_2(\theta^\star,\eta\rho_\epsilon)\big)$, 
 	and $\omega_{p,K} = \int_{p}^1 \sqrt{1-z^2}^Kdz$ if $K\geq0$ and $1$ else.
 }
\begin{remark}
The same result holds by replacing all occurrences of $f(\cdot)$ by the constant $f(t)$.
\end{remark}
\begin{remark}
In dimension $1$, the theorem takes a simpler form.
Indeed $C_{p,\eta,1}= 2$ for all $p,\eta\in(0,1)$ and thus, choosing $b=2$ for instance,  $C(1,2,\rho,p,\eta)$  reduces to $2\Big(2\max\Big\{\frac{2V_\rho^2}{\rho v^3_\rho} , \frac{V_\rho^{3/2}}{v_\rho^2},\frac{2V_\rho^{5/2}}{v_\rho^3} \Big\}
+1\Big)$. In  the case of  Bernoulli distributions,
if $\Theta_\rho = \{\log(\mu/(1-\mu)), \mu \in [\mu_\rho,1-\mu_\rho]\}$, then $v_\rho =\mu_\rho(1-\mu_\rho)$, $V_\rho=1/4$ and 
 $C(1,2,\rho,p,\eta) = 2(\frac{1}{8\mu_\rho^3(1-\mu_\rho)^3}+1) $.

\end{remark}
\begin{remark}	
	We believe  it is possible to  reduce the $\max$ term by a factor $V_\rho^3/v_\rho^4$
	 in the definition of $C(K,b,\rho,p,\eta)$.
\end{remark}

Let $f(x) = \log(x)+\xi\log\log(x)$.
We now state two corollaries of Theorem~\ref{thm:main},
The first one is stated for the case when boundary is set to $f(t)/n$ and is thus directly relevant to the analysis of \KLUCB. The second corollary is about the more challenging boundary $f(t/n)/n$ that corresponds to the \KLUCBp\ strategy. 
We note that $f$ is non-decreasing only for $x\geq e^{-\xi}$.
When $x=t$, this requires that $t\geq e^{-\xi}$.
Now, when $x=t/N_{a^\star}(t)$ where $N_{a\star}(t) = t-O(\ln(t))$,
imposing that $f$ is non-decreasing requires that $\xi\geq\ln(1-O(\ln(t)/t))$ for large $t$, that is $\xi\geq 0$.
In the sequel we thus restrict to $t\geq e^{-\xi}$ when using the boundary $f(t)$ and to $\xi\geq 0$ when using the boundary $f(t/n)$.
Finally, we remind that the quantity $\chi =  p\eta \sqrt{2v_\rho^2/V_\rho}$ is a function of $p,\eta$ and $\rho$, and introduce the notation $\chi_\epsilon = \rho_\epsilon\chi$ for convenience.

\corollarybox{Boundary crossing for $f(t)$ }{boundarycrossing}{
Let $f(x) = \log(x)+\xi\log\log(x)$. Using the same notations as in Theorem~\ref{thm:main}, for all $p,\eta\in[0,1], \rho<\rho^\star$
and all $t \geq e^{-\xi}$ such that $f(t)\geq 1$ it holds
\beqan
\lefteqn{
	\Pr_{\theta^\star}\Big\{\bigcup_{1\leq n < t}\hat \theta_n\in\Theta_\rho\cap \cK_{a^\star}(\Pi_{a^\star}(\hat \nu_{a^\star,n}),\mu^\star-\epsilon) \geq f(t)/n\Big\}}\\
&\leq&\frac{ C(K,4,\rho,p,\eta) (1+\chi_\epsilon)}{\chi_\epsilon t}
\bigg(1 +\xi\frac{\log\log(t)}{\log(t)}\bigg)^{K/2}\hspace{-2mm}\log(t)^{-\xi+K/2}e^{-\chi_\epsilon\sqrt{\log(t)+ \xi\log\log(t)}}\,.
\eeqan
}

\corollarybox{Boundary crossing  for $f(t/n)$ }{boundarycrossingplus}{
	
	Let $f(x) = \log(x)+\xi\log\log(x)$.
For all $p,\eta\in[0,1], \rho<\rho^\star$ and $\xi\geq \max(K/2-1,0)$, provided that $t\in[85 \chi^{-2},t_\chi]$
	where   
	$t_\chi =  \chi_\epsilon^{-2}\frac{\exp(\ln(4.5)^2/\chi_\epsilon^2)}{4\ln(4.5)^2}$, it holds
	\beqan
	\lefteqn{
		\Pr_{\theta^\star}\Big\{\bigcup_{1\leq n < t}\hat \theta_n\in\Theta_\rho\cap \cK_{a^\star}(\Pi_{a^\star}(\hat \nu_{a^\star,n}),\mu^\star-\epsilon) \geq f(t/n)/n\Big\} \leq C(K,4,\rho,p,\eta)\bigg[e^{-\chi_\epsilon\sqrt{t}c'} +}\\
	&&\hspace{-2mm}\frac{(1+\xi)^{K/2}}{ct \log(tc)}
	\begin{cases}
		\frac{16}{3}\log(tc\log(tc)/4)^{K/2-\xi}
		+80\log(1.25)^{K/2-\xi}&\text{if } \xi\geq K/2\\
		\frac{16}{3}\log(t/3)^{K/2-\xi}+ 80\log(t\frac{c\log(tc)}{4-c\log(tc)})^{K/2-\xi}&\text{if }  \xi\in[K/2-1,K/2]
	\end{cases} 
	\bigg],
	\eeqan
	where   $c = \chi_\epsilon^2/(2\log(5))^2$, and $c' = \sqrt{f(5)/5}$ if $\xi\geq K/2$ and $\sqrt{f(4)/4}$ else.
	Further, for larger values of $t$, $t\geq t_\chi$, the second term in the brackets becomes
	\beqan
	\frac{(1+\xi)^{K/2} }{ct \log(tc)}
	\begin{cases}
		144\log(1.25)^{K/2-\xi}&\text{if } \xi\geq K/2\\
		144\log(t/3)^{K/2-\xi}&\text{if }  \xi\in[K/2-1,K/2] \text{ (and } \xi\geq0)\,.
	\end{cases} 
	\eeqan
}
\begin{remark}
	In Corollary~\ref{cor:boundarycrossing}, since the asymptotic 
	regime of $\chi_\epsilon \sqrt{\log(t)} - (K/2-\xi)\log\log(t)$ may take a massive amount of time to kick-in when $\xi<K/2-2\chi_\epsilon$, we recommend to take  $\xi>K/2-2\chi_\epsilon$.
	Now, we also note that the value $\xi = K/2-1/2$ is interesting in practice, since then it holds $\log(t)^{K/2-\xi}
	=\sqrt{\log(t)}<5$ for all $t\leq 10^9$.
	\end{remark}
\begin{remark}	
	The restriction to $t\geq 85\chi_\epsilon^{-2}$ is merely for $\xi \simeq K/2-1$.
	For instance for $\xi\geq K/2$, the restriction becomes $t\geq 76\chi_\epsilon^{-2}$,
	and it becomes less restrictive for larger $\xi$.
	The term $t_\chi$ is virtually infinite: For instance when $\chi_\epsilon=0.3$, this is already larger than $10^{12}$, while $85\chi_\epsilon^{-2}<945$.
\end{remark}
\begin{remark}According to this result, the value $K/2-1$ (when it is non-negative) appears to be a critical value for $\xi$, since the boundary crossing probabilities are not summable in $t$ for $\xi\leq K/2-1$, but are summable for $\xi>K/2-1$.
	Indeed, the terms behind the curved brackets are conveniently $o(\log(t))$ with respect to $t$, except when $\xi=K/2-1$. In practice however, since this asymptotic behavior may take a large time to kick-in, we recommend  $\xi$ to be away from $K/2-1$.
\end{remark}

\begin{remark}Achieving a bound for  the threshold
	$f(t/N_a(t))$ is more challenging than for $f(t)$.
	Only the later case was analyzed in \cite{CaGaMaMuSt2013} as the former was 
	was out of reach of their analysis. Also, the result is valid with exponential families of dimension $K$ and not only dimension $1$, which is a major improvement. 
	It is interesting to note that when $K=1$, $\max(K/2-1,0)=0$, and to observe experimentally that a sharp phase transition indeed appears  for \KLUCBp\  precisely at the value $\xi=0$: the algorithm suffers a linear regret when $\xi<0$ and a logarithmic regret when $\xi=0$. For \KLUCB, no sharp phase transition appears at point $\xi=0$. Instead, a relatively smooth phase transition appears for a negative $\xi$ dependent on the problem. Both observations are coherent with the statements of the corollaries.
\end{remark}

{\bf Discussion regarding the proof technique}
The proof technique that we consider below significantly differs from the proof from \cite{CaGaMaMuSt2013} and \cite{HondaTakemura10DMED}, and combines 
key ideas disseminated in two works from Tze Leung Lai, \cite{lai1988boundary} and \cite{lai1987adaptive} with some non-trivial extension that we describe below.
Also, we also simplify sum of the original arguments and improve the readability of the initial proof technique, in order to shed more light on these neat ideas.

\noindent-{\bf Change of measure}
At a high level, the first main idea of this proof is to resort to a change of measure argument, which is the proof technique used to prove the lower bound on the regret. 
The work of  \cite{lai1988boundary} should be given full credit for this idea. 
This is in stark contrast with the proof techniques later developed for the finite-time analysis of stochastic bandits.
The change of measure is actually not used once, but twice. First, to go from $\theta^\star$, the parameter of the optimal arm
to some perturbation of it $\theta^\star_c$. Then, which is perhaps more surprising, to to go from this perturbed point to a mixture over a well-chosen ball centered on it.
Although we have reasons to believe that this second change of measure may not be  required (at least choosing a ball in dimension $K$ seems slightly sub-optimal),
this two-step localization procedure is definitely the first main component that enables to handle the boundary crossing probabilities.
The other steps for the proof of the Theorem include a concentration of measure argument and a peeling argument, which are more standard. 

\noindent-{\bf Bregman divergence}
The second main idea that is the use of Bregman divergence and its relation with the quadratic norm, which is due to \cite{lai1987adaptive}. This enables indeed to make explicit computations for exponential families of dimension $K$ without too much effort, at the price of loosing some "variance" terms (linked to the Hessian of the family).
We combine this idea with a some key properties of Bregman divergence that enables us to simplify a few steps, notably the concentration step, that we revisited entirely in order to obtain clean bounds valid in finite time and not only asymptotically.

\noindent-{\bf Concentration of measure and boundary effects}
One specific difficulty that appeared in the proof is to handle the shape of the parameter set $\Theta$, and the fact that $\theta^\star$ should be away from its boundary. The initial asymptotic proof of Lai did not account for this and was not entirely accurate. Going beyond this proved to be quite challenging due to the boundary effects, although the concentration result (section~\ref{sub:concentration}, Lemma~\ref{lem:concentrestricted}) that we obtain 
are eventually valid without restriction and the final proof looks deceptively easy.
This concentration result is novel.

\noindent-{\bf Cone covering and dimension $K$}
In \cite{lai1988boundary}, the author analyzed a boundary crossing problem
first in the case of exponential families of dimension $1$,
and then sketch the analysis for exponential families of dimension $K$ and 
for one the intersection with one cone. However the complete result was nowhere stated explicitly. As a matter of fact, the initial proof from \cite{lai1988boundary} restricts to a cone, which greatly simplifies the result. In order to obtain the full-blown results, valid in dimension $K$ for the unrestricted event, we introduced a cone covering of the space. This seemingly novel (although not very fancy) idea enables to get a final result that is only depending on the cone-covering number of the space. It required some careful considerations and simplifications of the initial steps from \cite{lai1988boundary}. Along the way, we made explicit the sketch of proof provided 
in \cite{lai1988boundary} for the dimension $K$.

\noindent-{\bf Corollaries and ratios}
The final key idea that should be credited to T.L. Lai is about the fine tuning of the final bound resulting from the two change of measures, the application of concentration and the peeling argument. Indeed these step lead to a bound by a sum of terms, say $\sum_{i=0}^I s_i$ that should be studied and depends on a few free parameters.
This correspond, with our rewriting and modifications, to the statement of Theorem~\ref{thm:main}. 

The brilliant idea of T.L. Lai, that we separate from the proof of Theorem~\ref{thm:main} and use in the proof of Corollaries~\ref{cor:boundarycrossing}
and \ref{cor:boundarycrossingplus} is to bound the ratios of $s_{i+1}/s_{i}$ for small values of $i$ and the ratio $s_{i}/s_{i+1}$ for large values of $i$ separately (instead of resorting, for instance to a sum-integral comparison lemma).
A careful study of these terms enable to improve the scaling and allow for smaller values of $\xi$, up to $K/2-1$, while other approaches seem unable to go below $K/2+1$.
Nevertheless, in our quest to obtain explicit bounds valid not-only asymptotically but also in finite time, this step is quite delicate, since a naive approach easily requires huge values for $t$ before the asymptotic regimes kick-in. By refining the initial proof strategy of \cite{lai1988boundary}, we managed to obtain a result valid for all $t$ for the setting of Corollary~\ref{cor:boundarycrossing}
and for all "reasonably"\footnote{We require $t$ to be at least about $10^2$ times some problem-dependent constant, against a factor that could be $e^{15}$ in the initial analysis.} large $t$ for the more challenging setting of Corollary~\ref{cor:boundarycrossingplus}.

\section{Analysis of boundary crossing probabilities: proof of Theorem~\ref{thm:main}}\label{sec:mainproof}
In this section, we closely follow the proof technique used in \cite{lai1988boundary} for the proof of Theorem~\ref{thm:mainLai},
in order to prove the result of Theorem~\ref{thm:main}. We precise further the constants,
remove the cone restriction on the parameter and modify the original proof to be fully non-asymptotic which, using the technique of \cite{lai1988boundary}, forces us to make some parts of the proof a little more accurate.

Let us recall that we consider $\Theta$ and $\rho$ such that $\theta^\star\in\Theta_{\rho} \subset \mathring{\Theta}_I$. The proof is divided in four main steps that we briefly present here for clarity:

In Section~\ref{sub:peeling}, we 
take care of the random number of pulls of the arm 
by a peeling argument. Simultaneously, we introduce a covering of the space with cones, which enables to later use arguments from 
proof of Theorem~\ref{thm:mainLai}.

In Section~\ref{sub:change}, we proceed with the first change of measure argument: taking advantage of the gap between $\mu^\star$ and $\mu^\star-\epsilon$, we move from a concentration argument around $\theta^\star$ to one around a shifted point $\theta^\star-\Delta_c$.

In Section~\ref{sub:localization}, we localize the 
empirical parameter $\hat \theta_n$ and make use of the second change of measure, this time to a mixture of measures, following \cite{lai1988boundary}. Even though we follow the same high level idea, we modified the original proof in order to better handle the cone covering,
and also make all quantities explicit.

In Section~\ref{sub:concentration}, we apply a concentration of measure argument. This part requires a specific care  since this is the core of the finite-time result. An important complication comes from the "boundary" of the parameter set, and was not explicitly controlled in the original proof from \cite{lai1988boundary}. A very careful analysis enables to obtain the finite-time concentration result without further restriction.

We finally combine all these steps in Sections~\ref{sub:combining}.

\subsection{Peeling and covering}\label{sub:peeling}
In this section, the intuition we follow is that 
we want to control the random number of pulls $N_{a^\star}(t) \in [1,t]$ and  to this end use a standard peeling argument, 
considering maximum concentration inequalities
on time intervals $[b^i,b^{i+1}]$ for some $b>1$.
Likewise, since the term $\cK_{a^\star}(\Pi_{a^\star}(\hat \nu_{a^\star,n}),\mu^\star-\epsilon) $ can be seen as an infimum of some quantity over the set of parameters $\Theta$, 
we use a covering of $\Theta$ in order to reduce the control of the desired quantity to that
 of each cell of the cover.
Formally, we show that 

\lemmabox{Peeling and cone covering decomposition}{peeling}{
For all $\beta\in(0,1),b>1$ and $\eta\in[0,1)$ it holds
\beqan
\lefteqn{
\Pr_{\theta^\star}\Big\{\bigcup_{1\leq n \leq t}\hat \theta_n\in\Theta_\rho\cap \cK_{a^\star}(\Pi_{a^\star}(\hat \nu_{a^\star,n}),\mu^\star-\epsilon) \geq f(t/n)/n\Big\}}\\
&\leq& \sum_{i=0}^{\lceil \log_b(\beta t+\beta)\rceil-2}
\sum_{c=1}^{C_{p,\eta,K}}\Pr_{\theta^\star}\Big\{\bigcup_{b^i\leq n < b^{i+1}} E_{c,p}(n,t)\Big\}
+
\sum_{c=1}^{C_{p,\eta,K}}\Pr_{\theta^\star}\Big\{\bigcup_{n=b^{\lceil \log_b(\beta t+\beta)\rceil-1}}^{t} E_{c,p}(n,t)\Big\}\,,
\eeqan
where the event $E_{c,p}(n,t)$ is defined by
\beqa
E_{c,p}(n,t)\eqdef \bigg\{ \hat \theta_n \in\Theta_\rho \cap 
\hat F_n \in\cC_p(\theta^\star_{c}) 
\cap \cB^\psi(\hat \theta_n, \theta^\star_{c}) \geq 
\frac{f(t/n)}{n}\bigg\}\,.
\eeqa
In this definition, $(\theta^\star_{c})_{c \leq C_{p,\eta,K}}$, constrained
to satisfy $\theta^\star_c \notin \cB_2(\theta^\star,\eta\rho_\epsilon)$, parameterize a minimal covering of $\nabla\psi(\Theta_\rho\setminus \cB_2(\theta^\star,\rho_\epsilon))$ with cones $\cC_p(\theta^\star_c) := \cC_{p}(\nabla\psi(\theta^\star_c);\theta^\star-\theta^\star_c)$
(That is $\nabla\psi(\Theta_\rho\setminus \cB_2(\theta^\star,\rho_\epsilon)) \subset\displaystyle{\bigcup_{c=1}^{C_{p,\eta,K}}\cC_p(\theta^\star_c)}$),
where $\cC_p(y;\Delta) = \bigg\{ y'\in\Real^K : \langle y'-y,\Delta\rangle \geq p \norm{y'-y} \norm{\Delta}\bigg\}$.
 For all $\eta<1$, $C_{p,\eta,K}$ is of order $(1-p)^{-K}$ and $C_{p,\eta,1}=2$, while $C_{p,\eta,K}\to\infty$ when $\eta\to1$. 
}

\medskip
{\bf Peeling}
Let us introduce an increasing sequence $\{n_i\}_{i\in \Nat}$ such that $n_0=1 < n_1 < \dots<n_{I_t}=t+1$ for some $I_t\in\Nat_\star$.
Then by a simple union bound it holds for any event $E_n$ 
\beqan
\Pr_{\theta^\star}\Big\{\bigcup_{1\leq n \leq t} E_n\Big\}
\leq \sum_{i=0}^{I_t-1}\Pr_{\theta^\star}\Big\{\bigcup_{n_i\leq n < n_{i+1}} E_n\Big\}\,.
\eeqan

We apply this simple result to the following sequence, defined for some $b>1$ and $\beta\in(0,1)$ by 
\beqan
n_i =
\begin{cases}
b^i  \text{ if } i < I_t \eqdef\lceil \log_b(\beta t+\beta)\rceil\\
t+1   \text{ if } i = I_t\,,
\end{cases}
\eeqan
(this is indeed a valid sequence since $n_{I_t-1} \leq b^{\log_b(\beta t+\beta)} = \beta (t+1) < t +1= n_{I_t}$), and to the event
\beqan
E_n \eqdef \bigg\{ \hat \theta_n\in\Theta_\rho\cap \cK_{a^\star}(\Pi_{a^\star}(\hat \nu_{a^\star,n}),\mu^\star-\epsilon) \geq f(t/n)/n\bigg\}\,.
\eeqan

{\bf Covering}
We now make the Kullback-Leibler projection explicit, and remark that in case of a regular family, it holds that
\beqan
\cK_{a^\star}(\Pi_{a^\star}(\hat \nu_{a^\star,n}),\mu^\star-\epsilon)
= \inf\bigg\{ \cB^\psi(\hat \theta_n, \theta^\star-\Delta) : \theta^\star - \Delta\in\Theta_\cD, \mu_{\theta^\star-\Delta} \geq \mu^\star-\epsilon \bigg\}\,,
\eeqan
where $\hat \theta_n\in\Theta_\cD$ is any point such that $\hat F_n = \nabla \psi(\hat \theta_n)$. 
This rewriting makes appear explicitly a shift from $\theta^\star$ to another point $\theta^\star-\Delta$.
For this reason, it is natural to study the link between
$\cB^\psi(\hat \theta_n, \theta^\star)$
and $\cB^\psi(\hat \theta_n, \theta^\star-\Delta)$.
Immediate computations show that  for any $\Delta$ such that $\theta^\star - \Delta\in\Theta_\cD$ it holds 
\beqa
\cB^\psi(\hat \theta_n,\theta^\star-\Delta)&=&
 \psi(\theta^\star-\Delta) - \psi(\hat \theta_n)
- \langle \theta^\star-\Delta-\hat \theta_n, \nabla \psi(\hat \theta_n)\rangle\nonumber\\
&=& 
\psi(\theta^\star) - \psi(\hat \theta_n)
- \langle \theta^\star-\hat \theta_n, \nabla \psi(\hat \theta_n)\rangle
+ \psi(\theta^\star-\Delta)-\psi(\theta^\star)
+\langle \Delta, \nabla \psi(\hat \theta_n)\rangle\nonumber\\
&=&\cB^\psi(\hat \theta_n,\theta^\star)
+ \psi(\theta^\star-\Delta)-\psi(\theta^\star)
+\langle \Delta, \nabla \psi(\hat \theta_n)\rangle\nonumber\\
&=&\cB^\psi(\hat \theta_n,\theta^\star) 
\underbrace{- \cB^\psi(\theta^\star-\Delta,\theta^\star)
-\langle \Delta,  \nabla\psi(\theta^\star-\Delta)-\hat F_n\rangle}_{\text{shift}}\,.\label{eqn:BregDecompo}
\eeqa
With this equality, the 
Kullback-Leibler projection can be rewritten to make appear an infimum over the shift term only.
In order to control the second part of the shift term we localize it thanks to a cone covering of $\nabla\psi(\Theta_\cD)$. 
More precisely, on the event $E_n$, we know that $\hat \theta_n \notin\cB_2(\theta^\star,\rho_\epsilon)$. Indeed, for all $\theta\in\cB_2(\theta^\star,\rho_\epsilon)\cap\Theta_\cD$, $\mu_\theta \geq \mu_\star-\epsilon$, and thus 
$\cK_{a^\star}(\nu_{\theta},\mu^\star-\epsilon)=0$.
It is thus natural to build a covering
of $\nabla\psi(\Theta_\rho\setminus\cB_2(\theta^\star,\rho_\epsilon))$. 
Formally, for a given  $p\in[0,1]$ and a base point $y\in\cY$, let us introduce the cone
\beqan
\cC_p(y;\Delta) = \bigg\{ y'\in\Real^K : \langle \Delta,y'-y\rangle \geq p  \norm{\Delta} \norm{y'-y}\bigg\}
\,.
\eeqan
We then associate to each $\theta \in\Theta_\rho$ a cone
 defined by $\cC_p(\theta) =\cC_p(\nabla\psi(\theta),\theta^\star-\theta)$.
Now for a given $p$, let $(\theta^\star_c)_{c=1,\dots,C_{p,\eta,K}}$ be a set of points corresponding to a minimal covering of the set $\nabla\psi(\Theta_\rho\setminus\cB_2(\theta^\star,\rho_\epsilon))$, in the sense that
\beqan
\nabla\psi(\Theta_\rho\setminus\cB_2(\theta^\star,\rho_\epsilon)) \subset \bigcup_{c=1}^{C_{p,\eta,K}} \cC_{p}(\theta^\star_c)\, \text{ with minimal } C_{p,\eta,K}\in\Nat\,,
\eeqan
constrained to be outside the ball $\cB_2(\theta^\star,\eta\rho_\epsilon)$, that is $\theta^\star_c \notin \cB_2(\theta^\star,\eta\rho_\epsilon)$ for each $c$.
It can be readily checked that by minimality of the size of the covering $C_{p,\eta,K}$, it must be that $\theta^\star_c \in\Theta_\rho\cap\cB_2(\theta^\star,\rho_\epsilon)$.
 More precisely, when $p<1$, then $\Delta_c = \theta^\star-\theta^\star_c$ is such that
$\rho_\epsilon- \norm{\Delta_c}$ is positive and away from $0$.
Also,  we have by property of $\cB_2(\theta^\star,\rho_\epsilon)$ that $\mu_{\theta^\star_c} \geq \mu^\star-\epsilon$, and by the constraint that $\norm{\Delta_c}>\eta\rho_\epsilon$.

  The size of the covering $C_{p,\eta,K}$ depends on the angle separation $p$, the ambient dimension $K$, and the repulsive parameter $\eta$. For instance it can be checked that $C_{p,\eta,1} = 2$ for all $p\in(0,1]$ and $\eta<1$. In higher dimension, $C_{p,\eta,K}$ typically scales as $(1-p)^{-K}$ and blows up when $p\to 1$. 
  It also blows up when $\eta\to 1$. 
It is now natural to introduce the decomposition
\beqa
E_{c,p}(n,t)\eqdef \bigg\{ \hat \theta_n \in\Theta_\rho \cap 
\hat F_n \in \cC_p(\theta^\star_c) 
\cap \cB^\psi(\hat \theta_n, \theta^\star_c) \geq 
\frac{f(t/n)}{n}\bigg\}\,.
\eeqa
Using this notation, we deduce that for all $\beta\in(0,1),b>1$
(we remind that $I_t= \lceil \log_b(\beta t+\beta)\rceil$),
\beqan
\Pr_{\theta^\star}\Big\{\bigcup_{1\leq n \leq t}\hat \theta_n\in\Theta_\rho\cap \cK_{a^\star}(\Pi_{a^\star}(\hat \nu_{a^\star,n}),\mu^\star-\epsilon) \geq f(t/n)/n\Big\}
&\leq&
\sum_{i=0}^{I_t-1}\sum_{c=1}^{C_{p,\eta,K}}\Pr_{\theta^\star}\Big\{\bigcup_{n_i\leq n < n_{i+1}} E_{c,p}(n,t)\Big\}\,.
\eeqan

\subsection{Change of measure}\label{sub:change}

In this section, we focus on one event $E_{c,p}(n,t)$. The  idea is to take advantage of the gap between $\mu^\star$ and $\mu^\star-\epsilon$,
that allows to shift from $\theta^\star$ to some of the $\theta^\star_c$ from the cover. The key observation
is to control the change of measure from $\theta^\star$ to each $\theta^\star_c$.
Note that $\theta^\star_c \in (\Theta_\rho \cap \cB_2(\theta^\star_c,\rho_\epsilon)) \setminus \cB_2(\theta^\star_c,\eta\rho_\epsilon)$ and that $\mu_{\theta^\star_c} \geq  \mu^\star-\epsilon$.
We show that

\lemmabox{Change of measure}{chgmeasure}{
If $n\to nf(t/n)$ is non-decreasing, then
for any  increasing sequence $\{ n_i \}_{i\geq 0}$ of non-negative integers 
it holds
\beqan
\Pr_{\theta^\star}\Bigl\{ \bigcup_{n=n_i}^{n_{i+1}-1}
E_{c,p}(n,t)\Bigr\}\leq
\exp\bigg(
- n_i\alpha^2
- \chi\sqrt{n_if(t/n_i)}\bigg)
\Pr_{\theta^\star_c}\Bigl\{ \bigcup_{n=n_i}^{n_{i+1}-1}
E_{c,p}(n,t)\Bigr\}
\eeqan
where $\alpha = \alpha(p,\eta,\epsilon) = \eta\rho_\epsilon\sqrt{v_\rho/2}$
and $\chi = p\eta\rho_\epsilon\sqrt{2v_\rho^2/V_\rho}$.
}
\begin{myproof}{of Lemma~\ref{lem:chgmeasure}}
For any event measurable $E$, we have by absolute continuity that
\beqan
\Pr_{\theta^\star}\Bigl\{ E \Bigr\} = \int_{E} \frac{d\Pr_{\theta^\star}}{d\Pr_{\theta^\star_c}}d\Pr_{\theta^\star_c}\,.
\eeqan
We thus bound the ratio which, in the case of $E = \{\bigcup_{n_i \leq n < n_{i+1}} E_{c,p}(n,t)\}$, leads to
\beqa
\int_{E}\frac{d\Pr_{\theta^\star}}{d\Pr_{\theta^\star_c}}d\Pr_{\theta^\star_c} &=&
\int_{E}\frac{\Pi_{k=1}^{n} \nu_{\theta^\star}(X_k)}{\Pi_{k=1}^{n} \nu_{\theta^\star_c}(X_k)}d\Pr_{\theta^\star_c}\nonumber\\
 &=& \int_{E}\exp\bigg( n\langle \theta^\star-\theta^\star_c, \hat F_{a^\star, n}\rangle  -  n\Big(\psi(\theta^\star) - \psi(\theta^\star_c)\Big)\bigg)d\Pr_{\theta^\star_c}\nonumber\\
 &=&
 \int_{E}\exp\bigg( -n \langle \Delta_c, \nabla \psi(\theta^\star_c) -\hat F_{a^\star, n}\rangle  -  n\cB^\psi(\theta^\star_c, \theta^\star)\bigg)d\Pr_{\theta^\star_c}\,,
 \label{eqn:ChangeM}
 \eeqa
 where $\Delta_c=  \theta^\star-\theta^\star_c$.
Note that this rewriting makes appear the same term
as the shift term appearing in \eqref{eqn:BregDecompo}. Now, we remark that since $\theta^\star_c\in\Theta_\rho$ by construction, then under the event 
 $E_{c,p}(n,t)$ it holds   by convexity of $\Theta_\rho$ and elementary Taylor approximation
 \beqa
 -\langle \Delta_c,  \nabla\psi(\theta^\star_c)-\hat F_n\rangle &\leq& -p  \norm{\Delta_c} ||\nabla\psi(\theta^\star_c)-\hat F_n||\nonumber\\
 &\leq& -p  \norm{\Delta_c}  v_\rho ||\hat \theta^\star_n - \theta^\star_c||\nonumber\\
 &\leq& -p \norm{\Delta_c} _c v_\rho\sqrt{\frac{2}{V_\rho}\cB^\psi(\hat \theta_n,\theta^\star_c)}\nonumber\\
 &\leq& -p \eta\rho_\epsilon v_\rho\sqrt{\frac{2f(t/n)}{V_\rho n} } \,.\label{eqn:BregDecomp2c}
\eeqa
where we used the fact that $ \norm{\Delta_c}  \geq \eta\rho_\epsilon$.
On the other hand, it also holds that 
\beqa
-\cB^\psi(\theta^\star_c,\theta^\star)
\leq -\frac{1}{2}v_\rho \norm{\Delta_c}^2 \leq 
-\frac{1}{2}v_\rho \eta^2\rho_\epsilon^2\,.\label{eqn:BregDecomp3c}
\eeqa
To conclude the proof we  plug-in \eqref{eqn:BregDecomp2c} and 
\eqref{eqn:BregDecomp3c} into \eqref{eqn:ChangeM}. Then, it remains to use that $n \geq b^i$ together with the fact that $n\mapsto nf(t/n)$ is non decreasing.
\end{myproof}

\subsection{Localized change of measure}\label{sub:localization}
In this section, we decompose further the event of interest in $\Pr_{\theta^\star_c}\Bigl\{ \bigcup_{n_i \leq n < n_{i+1}} E_{c,p}(n,t)\Bigr\}$
in order to apply some concentration of measure argument.
 In particular, since by construction
 \beqan
 \hat F_n \in \cC_p(\theta^\star_c) \Leftrightarrow
 \langle \Delta_c, \nabla \psi(\theta^\star_c) - \hat F_n\rangle
 \geq p \norm{\Delta_c} \norm{\nabla \psi(\theta^\star_c) - \hat F_n}\,,
 \eeqan
it is then natural to control 
$\norm{\nabla \psi(\theta^\star_c) - \hat F_n}$. This is what we call localization.
More precisely, we introduce 
for any sequence  $\{\epsilon_{t,i,c}\}_{t,i}$ of positive values, the following decomposition
\beqa
\Pr_{\theta^\star_c}\Bigl\{ \bigcup_{n_i \leq n < n_{i+1}} E_{c,p}(n,t)\Bigr\}
&\leq&
\Pr_{\theta^\star_c}\Bigl\{ \bigcup_{n_i \leq n < n_{i+1}} E_{c,p}(n,t)
\cap ||
\nabla \psi(\theta^\star_c) - \hat F_n|| < \epsilon_{t,i,c}\Bigr\}\nonumber\\
&&
+
\Pr_{\theta^\star_c}\Bigl\{ \bigcup_{n_i \leq n < n_{i+1}} E_{c,p}(n,t)
\cap ||
\nabla \psi(\theta^\star_c) - \hat F_n|| \geq \epsilon_{t,i,c}\Bigr\}\,.\label{eqn:DecompoEvents}
\eeqa
We handle the first term in \eqref{eqn:DecompoEvents} by another change of measure argument that we detail below, and the second term thanks to a concentration of measure argument that we detail in section~\ref{sub:concentration}.
We will show more precisely that

\lemmabox{Change of measure}{ChangeMesure}{
	 For any sequence of positive values $\{\epsilon_{t,i,c}\}_{i\geq 0}$, it holds
\beqan
\lefteqn{\Pr_{\theta^\star_c}\Big\{\bigcup_{n_i\leq n < n_{i+1}} E_{c,p}(n,t) \cap  \norm{\nabla \psi(\hat \theta_n)- \nabla \psi(\theta^\star_c)} < \epsilon_{t,i,c}\Big\}}\\
&\leq& \alpha_{\rho,p}\exp\Big(-f\Big(\frac{t}{n_{i+1}\!-\!1}\Big)\Big)\min\Big\{\rho^2v_\rho^2, \tilde\epsilon_{t,i,c}^2,\frac{(K+2)v_\rho^2}{K (n_{i+1}-1)V_{\rho}}\Big\}^{-K/2} \tilde\epsilon_{t,i,c}^{K}\,.
\eeqan
where $\tilde \epsilon_{t,i,c} = \min\{\epsilon_{t,i,c},\text{Diam}\big(\nabla\psi(\Theta_\rho) \cap \cC_p(\theta^\star_c)\big)\}$ and 
$\alpha_{\rho,p} = 2\frac{\omega_{p,K-2}}{\omega_{p',K-2}}
\bigg(\frac{V_\rho}{v_\rho^2}\bigg)^{K/2}\Big(\frac{V_\rho}{v_\rho}\Big)^K$ where
$p'>\max\{p,\frac{2}{\sqrt{5}}\}$, with $\omega_{p,K} = \int_{p}^1 \sqrt{1-z^2}^Kdz$ for $K\geq0$
and $w_{p,-1}=1$.
}

\medskip
Let us recall that 
$E_{c,p}(n,t) = \{\hat \theta_n\! \in\! \Theta_\rho\, \cap \hat F_n\! \in\! \cC_p(\theta^\star_c)\cap n\cB^{\psi}(\hat \theta_{n},\theta^\star_c) \geq f(t/n) \}$.

The idea is to go from $\theta^\star_c$ to the measure that corresponds to the mixture of all the
$\theta'$ in the shrink ball $B = \Theta_\rho \cap \nabla\psi^{-1}\big(\cC_p(\theta^\star_c) \cap \cB_2(\nabla\psi(\theta^\star_c),\epsilon_{t,i,c})\big)$ where
$\cB_2(y,r) \eqdef \Big\{ y'  \in \Real^K \,;\, \norm{y-y'}\leq  t \Big\}$.
This makes sense since, on the one hand, under $E_{c,p}(n,t)$,
$ \nabla\psi(\hat \theta_n) \in \cC_p(\theta^\star_c)$, and on the other hand, $||\nabla \psi(\hat \theta_n)- \nabla \psi(\theta^\star_c)|| \leq \epsilon_{t,i,c}$.
For convenience, let us
introduce the event of interest
\beqan
\Omega = \Big\{ \bigcup_{n_i\leq n < n_{i+1}}  \,\, E_{c,p}(n,t) \cap  \norm{\nabla \psi(\hat \theta_n)- \nabla \psi(\theta^\star_c)} \leq \epsilon_{t,i,c}\Big\}\,.
\eeqan
 We use the following change of measure
\beqan
d\Pr_{\theta^\star_c}=\frac{d\Pr_{\theta^\star_c}}{dQ_{B}}dQ_{B}\,,
\eeqan
where $Q_{B}(\Omega)  \eqdef \int_{\theta' \in B} \Pr_{\theta'}\Bigl\{\Omega\Bigr\}d\theta'$ is the mixture of all distributions with parameter in $B$.
The proof technique consists now in bounding the ratio by some quantity not depending on $\Omega$.

\beqan
\int_{\Omega}
\frac{d\Pr_{\theta}}{dQ_{B}}dQ_{B} &=&
\int_{\Omega} \Bigg[\int_{\theta' \in B}  \frac{\Pi_{k=1}^{n} \nu_{\theta'}(X_k)}{\Pi_{k=1}^{n} \nu_{\theta}(X_k)} d\theta'\Bigg]^{-1}dQ_{B}\\
&=& 
\int_\Omega\Bigg[\int_{\theta' \in B}  \exp\bigg(  n \langle\theta' - \theta,\hat F_{a^\star, n}\rangle -  n \big(\psi(\theta') - \psi(\theta)\big)\bigg) d\theta'\Bigg]^{-1}
dQ_{B}\,.
\eeqan
It is now convenient to remark that the term in the exponent can be rewritten in terms of Bregman divergence: by elementary substitution
of the definition of the divergence and of $\nabla \psi(\hat \theta_{n}) = \hat F_{a^\star, n}$, it holds
\beqan
\langle\theta' - \theta,\hat F_{a^\star, n}\rangle -  \big(\psi(\theta') - \psi(\theta)) = \cB^\psi(\hat \theta_n,\theta) - \cB^\psi(\hat \theta_n,\theta')\,.
\eeqan
Thus, the previous likelihood ratio simplifies as follows
\beqan
\frac{d\Pr_{\theta}}{dQ_{B}} &=& 
\Bigg[\int_{\theta' \in B}  \exp\bigg(n\cB^{\psi}(\hat \theta_{n},\theta) - n\cB^{\psi}(\hat \theta_{n},\theta')\bigg) d\theta'\Bigg]^{-1} \\
&\leq& \Bigg[\int_{\theta' \in B}  \exp\bigg(f(t/n) - n\cB^{\psi}(\hat \theta_{n},\theta') d\theta'\Bigg]^{-1}\\
&=& \exp\bigg(-f(t/n)\bigg) \Bigg[\int_{\theta'\in B} \exp\bigg(- n\cB^{\psi}(\hat \theta_{n},\theta')\bigg) dx\Bigg]^{-1}\,,
\eeqan
where we we note that both $\theta'$ and $\hat \theta_{n}$ belong to $\Theta_{\rho}$.

The next step is to consider a set $B' \subset B$ that contains $\hat \theta_n$. For each such set, 
and the upper bound $\cB^\psi(\hat \theta_n,\theta') \leq \frac{V_\rho}{2v_\rho^2}\norm{\nabla\psi(\hat \theta_n)- \nabla\psi(\theta')}^2$,
we now obtain 
\beqan
\frac{d\Pr_{\theta}}{dQ_{B}} 
&\stackrel{(a)}{\leq}&  \exp\bigg(-f(t/n)\bigg)
\Bigg[\int_{\theta'\in  B'} \exp\bigg(- \frac{nV_\rho}{2v_\rho^2} \norm{\nabla\psi(\hat \theta_n)- \nabla\psi(\theta')}^2 \bigg) d\theta'\Bigg]^{-1}\\
&\stackrel{(b)}{=}& \exp\bigg(-f(t/n)\bigg)
\Bigg[\int_{y\in 	 \nabla \psi(B')}\! \exp\bigg(\!-\! \frac{nV_\rho}{2v_\rho^2} \norm{\nabla \psi (\hat \theta_n)\!-\!y}^2 \bigg) |\det(\nabla^2\psi^{-1}(y))|dy\Bigg]^{-1}\\
&\stackrel{(c)}{\leq}&\exp\bigg(-f(t/n)\bigg)
\Bigg[\int_{y\in 
	\nabla \psi(B')}\! \exp\bigg(\!-\! \frac{nV_\rho}{2v_\rho^2} \norm{\nabla \psi (\hat \theta_n)\!-\!y}^2 \bigg) dy\Bigg]^{-1}\!V_\rho^{K}\,.
\eeqan
In this derivation, $(a)$ holds by positivity of $\exp$ and the inclusion $B'\subset B$,
$(b)$ follows by a change of parameter argument and $(c)$ is obtained by controlling the determinant (in dimension $K$)
of the Hessian, whose highest eigenvalue is $V_\rho$.

In order to identify a good candidate for the set $B'$ let us now study the set $B$.
A first remark is that $\theta^\star_c$ plays a central role in $B$:  It is not difficult to show that, by construction of $B$,
\beqan
\nabla\psi^{-1}\bigg( \nabla\psi(\theta^\star_c) + 
\cB_2(0,\min\{v_\rho \rho,\epsilon_{t,i,c}\})\cap \cC_p(0;\Delta_c) \bigg) \subset B\,.
\eeqan
Indeed,  if $\theta'$ belongs to the set on the left hand side, then it must satisfy  on the one hand$\nabla \psi(\theta') \in \nabla\psi(\theta^\star_c) + \cB_2(0,v_\rho \rho)$. This implies that $\theta'\in \cB_2(\theta^\star_c,\rho)\subset \Theta_\rho$ (this last inclusion is by construction of $\Theta$).
On the other hand, it satisfies $\nabla \psi(\theta') \in \nabla \psi(\theta^\star_c)+\cB_2(0,\epsilon_{t,i,c})\cap \cC_p(0,\Delta_c)$.
These two properties show that such a $\theta'$ belongs to $B$.

Thus, a natural candidate $B'$ should satisfy
$\nabla \psi(\!B'\!) \subset  \nabla \psi(\theta^\star_c) + \cB_2(0,\tilde r)\!\cap\! \cC_{p}(0;\!\Delta_c)$,
with $\tilde r =\! \min\{\!v_\rho \rho,\epsilon_{t,i,c}\!\}$.
It is then natural to look for $B'$ in the form
$\nabla \psi^{-1}(\nabla \psi(\theta^\star_c) + \cB_2(0,\tilde r)\cap \cD)$,
where  $\cD \subset \cC_{p}(0;\Delta_c)$ is a sub-cone of $\cC_{p}(0;\Delta_c)$ with base point $0$.  In this case, the previous derivation simplifies into
\beqan
\frac{d\Pr_{\theta}}{dQ_{B}} 
&\leq&\exp\bigg(-f(t/n)\bigg)
\Bigg[\int_{y\in \cB_2(0,\tilde r)\cap\cD}\! \exp\bigg(\!-\! C \norm{y_n\!-\!y}^2 \bigg) dy\Bigg]^{-1}\!V_\rho^{K}\,,
\eeqan
where  $y_n = \nabla \psi (\hat \theta_n)-\nabla \psi(\theta^\star_c) \in \cB_2(0,\tilde r)\cap\cD$ and $C= \frac{nV_\rho}{2v_\rho^2}$.
Cases of special interest for the set $\cD$ are such that the value of the function  $g:y \mapsto \int_{y'\in \cB_2(0,\tilde r)\cap \cD} \exp\big(- C\norm{y-y'}^2\big)dy'$, for $y\in\cB_2(0,\tilde r)\cap \cD$  is minimal at the base point $0$. Indeed this enables to derive the following bound
\beqan
\frac{d\Pr_{\theta}}{dQ_{B}}
&\leq &\exp\bigg(-f(t/n)\bigg)
\Bigg[\int_{y\in 
	\cB_2(0,\min\{v_\rho \rho,\epsilon_{t,i,c}\})\cap \cD} \exp\bigg(- \frac{nV_\rho}{2v_\rho^2} \norm{y}^2 \bigg) dy\Bigg]^{-1}V_\rho^{K}\\
&\stackrel{(d)}{=}& \exp\bigg(-f(t/n)\bigg)
\Bigg[\int_{y\in   
	\cB_2(0,r_{\rho})\cap \cD} \exp\bigg(- n\norm{y}^2 \bigg) dy\Bigg]^{-1} \bigg(\frac{V_\rho^2}{2v_\rho^2}\bigg)^K\,,
\eeqan
where
$(d)$ follows from another change of parameter argument, 
with $r_\rho= \sqrt{\frac{V_\rho}{2v_\rho^2}}\min\{v_\rho \rho,\epsilon_{t,i,c}\}$ combined with isotropy of the Euclidean norm (the right hand side of $(d)$ no longer depends on the random direction $\Delta_n$), plus the fact that the sub-cone
$\cD$ is invariant by rescaling. We recognize here a Gaussian integral on $\cB_2(0,r_{\rho})\cap \cD$ that can be bounded explicitly (see below).

Following this reasoning, we are now ready to specify the set $\cD$.
Let $\cD = \cC_{p'}(0;\Delta_n) \subset\cC_p(0;\Delta_c)$ be a sub-cone where  $p'\geq p$ (remember that the larger $p$, the more acute is a cone) and $\Delta_n$ is chosen such that $\nabla \psi(\hat \theta_n) \in \nabla \psi(\theta^\star_c)+ \cD$ (there always exists such a cone). It thus remains to specify $p'$.
A study of the function $g$ (defined above) on the domain $\cB_2(0,\tilde r)\cap \cC_{p'}(0;\Delta_n)$ reveals that it is minimal at point $0$ provided that $p'$ is not too small, more precisely provided that $p \geq 2/\sqrt{5}$.
The intuitive reasons are that the points that contribute most to the integral belong to the set $\cB_2(y,r)\cap \cB_2(0,\tilde r)\cap\cD$ for small values of $r$, that this set has lowest volume (the map $y \to |\cB_2(y,r)\cap \cB_2(0,\tilde r)\cap\cD|$ is minimal) when $y\in  \partial\cB_2(0,\tilde r)\cap\partial\cD$ and that $y=0$ is a minimizer amongst these point provided that $p'$ is not too small. More formally,  the function $g$ rewrites
\beqan
g(y) &=&  \int_{r=0}^\infty e^{-Cr^2}| \cS_2(y,r) \cap \cB_2(0,\tilde r)\cap\cD|dr\,,
\eeqan
from which we see that a minimal $y$ should be such that 
the spherical section
$| \cS_2(y,r) \cap \cB_2(0,\tilde r)\cap\cD|$ is minimal for small values of $r$ (note also that $C= O(n)$). 
Then, since $B=\cB_2(0,\tilde r)\cap\cD$ is a convex set, the sections 
$| \cS_2(y,r) \cap \cB_2(0,\tilde r)\cap\cD|$ are of minimal size for points $y\in B$ that are extremal, in the sense that $y$ satisfies
$B \subset \cB_2(y,\text{Diam}(B))$. 
In order to choose $p'$ and fully specify $\cD$, we finally use the following lemma:
\begin{lemma}\label{lem:conesballs}
	Let $\cC_{p'}= \{ y' : \langle y', \Delta\rangle \geq p' \norm{y'}\norm{\Delta} \}$
	be a cone with base point $0$ and define $B = \cB_2(0,r)\cap \cC_{p'}$.
	Provided that  $p'> 2/\sqrt{5}$, then the set of extremal points 
$\{ y\in B: B \subset \cB_2(y,\text{Diam}(B))\}$ reduces to $\{0\}$.
\end{lemma}
\begin{myproof}{of Lemma~\ref{lem:conesballs}}
First, note that the boundary of the convex set $B$
is supported by the union of the base point $0$ 
and the set $\partial \cB_2(0,\tilde r) \cap \partial \cD$. Since this set is a sphere in dimension $K-1$ with radius  $\frac{\sqrt{1-{p'}^2}}{p}\tilde r$, 
all its points are at distance at most $2\frac{\sqrt{1-{p'}^2}}{p'}\tilde r$ from each other.
Now they are also at distance exactly $\tilde r$ from the base point $0$.
Thus, when $2\frac{\sqrt{1-{p'}^2}}{p'}\tilde r <\tilde r$, that is
$p'> 2/\sqrt{5}$, then $0$ is the unique point that satisfies
$B \subset \cB_2(y,\text{Diam}(B))$.
\end{myproof}
We now summarize the previous steps. So far, we have proved the following upper bound
\beqan
\Pr_{\theta^\star_c}\Bigl\{ \Omega \Bigr\}\!
&\!\leq\!&
\!\!\max_{n_i\leq n< n_{i+1}}\!
\exp\bigg(\!-\!f(t/n)\bigg)
\Bigg[\int_{y\in   
	\cB_2(0,r_\rho)\cap \cC_{p'}(0;{\bf 1})}\! \exp\big(\!-\!n\norm{y}^2 \big) dy\Bigg]^{-1} \!\bigg(\frac{ V_\rho^2}{2v_\rho^2}\bigg)^K\!\int_{\theta' \in B}\!\Pr_{\theta'}\Bigl\{ \!\Omega\! \Bigr\}d\theta'\\
&\leq&\!
\exp\bigg(\!-\!f(t/(n_{i+1}\!-\!1))\bigg)
\Bigg[\int_{y\in   
	\cB_2(0,r_\rho)\cap \cC_{p'}(0;{\bf 1})}\! \exp\big(\!-\!(n_{i+1}\!-\!1)\norm{y}^2 \big) dy\Bigg]^{-1}\! \bigg(\frac{ V_\rho}{2v_\rho^2}\bigg)^K\! V_\rho^K|B|\,.
\eeqan
where $|B|$ denotes the volume of $B$, $r_\rho= \sqrt{\frac{V_\rho}{2v_\rho^2}}\min\{v_\rho \rho,\epsilon_{t,i,c}\}$
and  for $p'>\max\{p,2/\sqrt{5}\}$,  
We remark that by definition of $B$, it holds
\beqan
|B| &\leq& \sup_{\theta\in\Theta_\rho} \det(\nabla^2\psi^{-1}(\theta)) |\cB_2(0,\epsilon_{t,i,c})\cap \cC_p(0;{\bf 1})|\\
&\leq& v_\rho^{-K}|\cB_2(0,\epsilon_{t,i,c})\cap \cC_p(0;{\bf 1})|\,.
\eeqan

Thus, it remains to analyze the 
volume and the Gaussian integral of 
$\cB_2(0,\epsilon_{t,i,c})\cap \cC_p(0;{\bf 1})$.
To do so, we use the following result from elementary geometry,
whose proof is given in Appendix~\ref{app}:
\begin{lemma}\label{lem:volumecones}
For all $\epsilon,\epsilon'>0$, $p,p'\in[0,1]$ and all $K\geq 1$ the following equality and inequality hold
	\beqan
	\frac{
		|\cB_2(0,\epsilon) \cap \cC_p(0;{\bf 1})|}{\int_{\cB_2(0,\epsilon') \cap \cC_{p'}(0;{\bf 1})}e^{- \norm{y}^2/2} dy }&=& \frac{\omega_{p,K-2}}{\omega_{p',K-2}}\frac{\int_{0}^\epsilon r^{K-1} dr}{\int_{0}^{\epsilon'} e^{-r^2/2} r^{K-1}dr}
	\leq
2\frac{\omega_{p,K-2}}{\omega_{p',K-2}} \bigg(\frac{\epsilon}{ \min \{ \epsilon',\sqrt{1+2/K}\}}\bigg)^K\,,
	\eeqan
	where  $\omega_{p,K-2}=\int_{p}^1 \sqrt{1-z^2}^{K-2}dz$ for $K\geq 2$ and  
	using the convention that $\omega_{p,-1}=1$.
	
\end{lemma}

Applying this Lemma, we thus get for  $r_\rho= \sqrt{\frac{V_\rho}{2v_\rho^2}}\min\{v_\rho \rho,\epsilon_{t,i,c}\}$,
\beqan
\Pr_{\theta^\star_c}\Bigl\{ \Omega \Bigr\}
&\leq&
e^{-f\Big(\frac{t}{n_{i+1}\!-\!1}\Big)}
\frac{\bigg(\frac{ V_\rho}{2v_\rho^2}\bigg)^K \Big(\frac{V_\rho}{v_\rho}\Big)^K |\cB_2(0,\epsilon_{t,i,c})\cap \cC_{p}(0;{\bf 1})|}{
\int_{y\in   
	\cB_2(0,r_\rho)\cap \cC_{p'}(0;{\bf 1})} \exp\big(\!-\!(n_{i+1}\!-\!1)\norm{y}^2 \big) dy} \\
&=&e^{-f\Big(\frac{t}{n_{i+1}\!-\!1}\Big)}
\bigg(\frac{ V_\rho}{v_\rho^2}\bigg)^K \Big(\frac{V_\rho}{v_\rho}\Big)^K(n_{i+1}\!-\!1)^{K/2}
\frac{|\cB_2(0,\epsilon_{t,i,c})\cap \cC_{p}(0;{\bf 1})|}
{\int_{y\in   
		\cB_2(0,\sqrt{2(n_{i+1}\!-\!1)}r_\rho)\cap \cC_{p'}(0;{\bf 1})} \exp\big(\!-\!\norm{y}^2/2 \big) dy}\\
	&\leq&2\frac{\omega_{p,K-2}}{\omega_{p',K-2}} e^{-f\Big(\frac{t}{n_{i+1}\!-\!1}\Big)}
	\bigg(\frac{ V_\rho}{v_\rho^2}\bigg)^K\Big(\frac{V_\rho}{v_\rho}\Big)^K (n_{i+1}\!-\!1)^{K/2}
	\Bigg(\frac{\epsilon_{t,i,c}}{\min\{\sqrt{2(n_{i+1}\!-\!1)}r_\rho, \sqrt{1+2/K}\}}\Bigg)^K\\
	&=&2\frac{\omega_{p,K-2}}{\omega_{p',K-2}} e^{-f\Big(\frac{t}{n_{i+1}\!-\!1}\Big)}
	\bigg(\frac{ V_\rho}{v_\rho^2}\bigg)^K\Big(\frac{V_\rho}{v_\rho}\Big)^K
	\Bigg(\frac{\epsilon_{t,i,c}^2}{\min\{  v_\rho^2 \rho^2, \epsilon_{t,i,c}^2, \frac{(K+2)v_\rho^2}{KV_\rho (n_{i+1}\!-\!1)}\}}\Bigg)^{K/2}\Big(\frac{V_\rho}{v_\rho^2}\Big)^{-K/2}\\
		&=&2\frac{\omega_{p,K-2}}{\omega_{p',K-2}}
	\bigg(\frac{V_\rho}{v_\rho^2}\bigg)^{K/2}\Big(\frac{V_\rho}{v_\rho}\Big)^K e^{-f\Big(\frac{t}{n_{i+1}\!-\!1}\Big)}
	\min\Big\{  v_\rho^2 \rho^2, \epsilon_{t,i,c}^2, \frac{(K+2)v_\rho^2}{KV_\rho (n_{i+1}\!-\!1)}\Big\}^{-K/2}\epsilon_{t,i,c}^K\,.
\eeqan
This concludes the proof of Lemma~\ref{lem:ChangeMesure}.

\subsection{Concentration of measure}\label{sub:concentration}
In this section, we focus on the second term in
\eqref{eqn:DecompoEvents}, that is we want to control
$\Pr_{\theta^\star_c}\Bigl\{ \bigcup_{n_i \leq n < n_{i+1}} E_{c,p}(n,t)
\cap ||
\nabla \psi(\theta^\star_c) - \hat F_n|| \geq \epsilon_{t,i,c}\Bigr\}$. In this term, $\epsilon_{t,i,c}$
should be considered as decreasing fast to $0$ with $i$, and slowly increasing with $t$. 
Note that by definition $\nabla \psi(\hat \theta_n) = \hat F_{a^\star, n}  = \frac{1}{n}\sum_{i=1}^n F(X_{a^\star,i}) \in\Real^K$
is an empirical mean with mean given by $\nabla \psi(\theta^\star_c)\in\Real^K$ and covariance matrix $\frac{1}{n}\nabla^2 \psi(\theta^\star_c)$. We thus resort to 
a concentration of measure argument.
 
 \lemmabox{Concentration of measure}{concentrestricted}{
Let $\epsilon^{\max}_c = 
\text{Diam}(\nabla \psi(\Theta_\rho\! \cap\! \cC_{c,p}))$
where we introduced the projected cone $\cC_{c,p} = \{\theta \!\in\!\Theta :  \langle\frac{\Delta_c}{\norm{\Delta_c}}, \frac{\nabla \psi(\theta^\star_c)-\nabla \psi(\theta)}{||
	\nabla \psi(\theta^\star_c)-\nabla \psi(\theta)||}\rangle \geq p \}$.
Then, for all $\epsilon_{t,i,c}$, it holds
 \beqan
 \Pr_{\theta^\star_c}\Big\{\!\bigcup_{n=n_i}^{n_{i+1}-1}\!\!E_{c,p}(n,t) \cap||\nabla \psi(\hat \theta_n)\!-\! \nabla \psi(\theta^\star_c)||\! \geq\! \epsilon_{t,i,c}\Big\}\!
\leq \exp\!\bigg(\!-\!\frac{n_i^2p\epsilon_{t,i,c}^2}{2V_{\rho} (n_{i+1}\!-\!1)}\bigg)\indic{\epsilon_{t,i,c}\!\leq\!\overline{\epsilon}_c}
 .
 \eeqan
 }

 
\begin{myproof}{of Lemma~\ref{lem:concentrestricted}}
Note that  by definition if $\epsilon_{t,i,c} > 	 \epsilon^{\max}_c $, then
\beqan	
\Pr_{\theta^\star_c}\Big\{\bigcup_{n_i\leq n < n_{i+1}} 
E_{c,p}(n,t) \cap  ||\nabla \psi(\hat \theta_n)- \nabla \psi(\theta^\star_c)|| \geq \epsilon_{t,i,c}\Big\}=0\,.
\eeqan
We thus restrict to the case when 
$\epsilon_{t,i,c}\leq  \epsilon^{\max}_c$, or equivalently,
replace $\epsilon_{t,i,c}$
by $\tilde \epsilon_{t,i,c} = \min \{\epsilon_{t,i,c},\epsilon^{\max}_c\}$. Now, by definition of the event $E_{c,p}(n,t)$, we have the rewriting
	\beqan
	\lefteqn{
		\Pr_{\theta^\star_c}\Big\{\bigcup_{n_i\leq n < n_{i+1}} 
		E_{c,p}(n,t) \cap  ||\nabla \psi(\hat \theta_n)- \nabla \psi(\theta^\star_c)|| \geq \tilde \epsilon_{t,i,c}\Big\}}\\
	\!&\!\leq\!&\!\!
			\Pr_{\theta^\star_c}\Big\{\bigcup_{n_i\leq n < n_{i+1}} 
\hat \theta_n \in\Theta_\rho  \cap  \langle \frac{\Delta_{c}}{\norm{\Delta_c}}, \nabla \psi(\theta^\star_c)-\nabla \psi(\hat \theta_n)\rangle\geq
p\tilde \epsilon_{t,i,c}\Big\}\\
	\!&\!\leq\!&\!\!
\Pr_{\theta^\star_c}\Big\{\bigcup_{n=n_i}^{n_{i+1}-1} \langle \frac{\Delta_{c}}{\norm{\Delta_c}}, \sum_{i=1}^n\big(\nabla \psi(\theta^\star_c)-F(X_{a^\star,i})\big)\rangle\geq
pn_i\tilde \epsilon_{t,i,c}\Big\}\,.
\eeqan
Now, applying on both side of the inequality the function $x\mapsto\exp(\lambda x)$, for a  deterministic $\lambda>0$, it comes
\beqan
	\lefteqn{
	\Pr_{\theta^\star_c}\Big\{\bigcup_{n_i\leq n < n_{i+1}} 
	E_{c,p}(n,t) \cap  ||\nabla \psi(\hat \theta_n)- \nabla \psi(\theta^\star_c)|| \geq \tilde \epsilon_{t,i,c}\Big\}}\\
\!&\!\stackrel{(a)}{\leq}\!&\!\!
\Pr_{\theta^\star_c}\Big\{\bigcup_{n=n_i}^{n_{i+1}-1}
\exp\bigg(\sum_{i=1}^n\langle \frac{\lambda \Delta_{c}}{\norm{\Delta_c}}, \nabla \psi(\theta^\star_c)-F(X_{a^\star,i})\rangle\bigg)\geq
\exp\Big(\lambda pn_i\tilde \epsilon_{t,i,c}\Big)\Big\}\\
	\!&\!=\!&\!\!
\Pr_{\theta^\star_c}\Big\{\!\!\bigcup_{n=n_i}^{n_{i+1}-1}\!\!\!
\exp\!\bigg(\!\sum_{i=1}^n\langle\! \frac{\lambda \Delta_{c}}{\norm{\Delta_c}}, \nabla \psi(\theta^\star_c)\!-\!F(X_{a^\star,i})\!\rangle \!-\! \frac{\lambda^2(n_{i+1}\!\!-\!\!1)}{2} V_\rho\!\bigg)\!\!\geq\!
\exp\!\Big(\lambda pn_i\tilde \epsilon_{t,i,c}\!-\!\frac{\lambda^2(n_{i+1}\!\!-\!\!1)}{2} V_\rho\Big)\!\Big\}\\
	\!&\!\leq\!&\!\!
\Pr_{\theta^\star_c}\Big\{\bigcup_{n=n_i}^{n_{i+1}-1}
\exp\bigg(\sum_{i=1}^n\langle \frac{\lambda \Delta_{c}}{\norm{\Delta_c}}, \nabla \psi(\theta^\star_c)\!-\!F(X_{a^\star,i})\rangle - \frac{\lambda^2n}{2} V_\rho\bigg)\geq
\exp\Big(\lambda pn_i\tilde \epsilon_{t,i,c}\!-\!\frac{\lambda^2(n_{i+1}\!\!-\!\!1)}{2} V_\rho\Big)\Big\}\,.
	\eeqan
	Now we recognize that the sequence $\{W_n(\lambda)\}_{n\geq 0}$, where $W_n(\lambda) = \exp\bigg(\sum_{i=1}^n\langle \frac{\lambda \Delta_{c}}{\norm{\Delta_c}}, \nabla \psi(\theta^\star_c) -F(X_{a^\star,i})\rangle - n\frac{\lambda^2 V_\rho}{2}\big)\bigg)$ is a non-negative super-martingale provided that $\lambda$ is not too large. 
	Indeed, provided that $\theta^\star_c -\frac{\lambda \Delta_{c}}{\norm{\Delta_c}}\in\Theta_{\rho}$ it holds
	\beqan
	\lefteqn{
		\Esp_{\theta^\star_c}\bigg[\exp\bigg(\sum_{i=1}^n\lambda \langle \frac{\Delta_{c}}{\norm{\Delta_c}}, \nabla \psi(\theta^\star_c) -F(X_{a^\star,i})\rangle - \frac{\lambda^2n V_\rho}{2}\big)\bigg) \bigg| \cH_{n-1} \bigg] }\\
	&\leq&\exp\bigg(\sum_{i=1}^{n-1}\lambda \langle \frac{\Delta_{c}}{\norm{\Delta_c}}, \nabla \psi(\theta^\star_c)
\!-\!	F(X_{a^\star,i})\rangle\!-\! (n\!-\!1)\frac{\lambda^2 V_\rho}{2}\bigg)\\
	&&\times
	\underbrace{\Esp_{\theta^\star_c}\bigg[\exp\bigg(\lambda \langle \frac{\Delta_{c}}{\norm{\Delta_c}},\nabla \psi(\theta^\star_c) \!-\!
		 F(X_{a^\star,n})\rangle\!-\! \frac{\lambda^2 V_\rho}{2}\bigg)\bigg| \cH_{n-1}\bigg]}_{\leq 1}
	\,,
	\eeqan
	that is $\Esp\bigg[W_n(e,\lambda) \bigg| H_{n-1}\bigg] \leq W_{n-1}(e,\lambda)$. 
	Thus, we apply Doob's maximal inequality for non-negative super-martingale and deduce that
	\beqan
	\lefteqn{
		\Pr_{\theta^\star_c}\Big\{\bigcup_{n_i\leq n < n_{i+1}} \,\,		E_{c,p}(n,t) \cap  ||\nabla \psi(\hat \theta_n)- \nabla \psi(\theta^\star_c)|| \geq \tilde \epsilon_{t,i,c}\Big\}}\\
	&\leq&
	\Pr_{\theta^\star_c}\Big\{\max_{n_i\leq n <n_{i+1}} \,\, W_n(\lambda)\geq \exp\bigg(\lambda pn_i\tilde \epsilon_{t,i,c}-\lambda^2(n_{i+1}\!-\!1)V_\rho/2\bigg)\Big\}\\
	&\leq&
	\Esp_{\theta^\star_c}[W_{n_i}(\lambda)]\exp\bigg(-\lambda pn_i\tilde \epsilon_{t,i,c} +\lambda^2(n_{i+1}\!-\!1)V_\rho/2\bigg)\\
	&\leq&
	\exp\bigg(-\lambda p n_i\tilde \epsilon_{t,i,c} +\lambda^2(n_{i+1}\!-\!1)V_\rho/2\bigg)\,.
	\eeqan
	Optimizing over $\lambda$ gives $\lambda =
	\lambda^\star= 
	\frac{n_ip\tilde \epsilon_{t,i,c}}{(n_{i+1}\!-\!1)V_\rho}$, and thus the condition becomes
	$\theta^\star_c - \frac{n_ip\tilde \epsilon_{t,i,c}}{(n_{i+1}\!-\!1)V_\rho\norm{\Delta_c}} \Delta_c \in\Theta_{\rho}$.
	At this point, it is convenient to introduce the quantity
	\beqan
	\lambda_c = \argmax\{ \lambda : \theta^\star_c - \lambda \frac{\Delta_c}{\norm{\Delta_c}} \in \Theta_\rho \cap \cC_{c,p} \}\,.
	\eeqan
	Indeed, it suffices to show that $\lambda^\star\leq\lambda_{c}$ to ensure that the condition is satisfied. It is now not difficult to relate $\lambda_c$ to $\epsilon^{\max}_{c}$:
	Indeed, any $\theta_\lambda = \theta^\star_c - \lambda \frac{\Delta_c}{\norm{\Delta_c}}$ that maximizes $||\nabla \psi(\theta^\star_c)-\nabla \psi(\theta_\lambda)||$ and belongs to $\Theta_\rho$ must satisfy
	\beqan
	\langle\frac{\Delta_c}{\norm{\Delta_c}},\nabla \psi(\theta^\star_c)-\nabla \psi(\theta_\lambda) \rangle
	 &\geq& p \overline{\epsilon}_c
	\eeqan
	on the one hand, and on the other hand, since $\theta^\star_c,\theta_\lambda \in\Theta_\rho$,
\beqan
	\langle\frac{\Delta_c}{\norm{\Delta_c}},\nabla \psi(\theta^\star_c)-\nabla \psi(\theta_\lambda) \rangle
	\leq V_{\rho} \norm{\frac{\Delta_c}{\norm{\Delta_c}}}
	\norm{\theta^\star_c-\theta_\lambda} = V_{\rho} \lambda\,.
\eeqan
Combining these two inequalities, we deduce that $\lambda_c \geq  p \epsilon^{\max}_{c}/V_{\rho}$.
Thus, using that $n_{i}/(n_{i+1}\!-\!1)\leq 1$ and 
$\tilde \epsilon_{t,i,c}\leq  \epsilon^{\max}_{c}$, we deduce that
$\lambda^\star = 	\frac{n_ip\tilde \epsilon_{t,i,c}}{(n_{i+1}\!-\!1)V_{\rho}} \leq
	\frac{p \overline{\epsilon}_c}{V_\rho} \leq \lambda_{c}$ is indeed satisfied.
		We then get	without further restriction
	\beqa
	\Pr_{\theta^\star_c}\Big\{\bigcup_{n_i\leq n < n_{i+1}}\!E_{c,p}(n,t) \cap||\nabla \psi(\hat \theta_n)\!-\! \nabla \psi(\theta^\star_c)|| \geq \epsilon_{t,i,c}\Big\}
	\leq \exp\bigg(-\frac{n_i^2p\epsilon_{t,i,c}^2}{2V_\rho (n_{i+1}\!-\!1)}\bigg)\indic{\epsilon_{t,i,c}\leq\overline{\epsilon}_c}
	\,.\label{eqn:concentbi}
	\eeqa
	
\end{myproof}

\subsection{Combining the different steps}\label{sub:combining}
In this part, we recap what we have shown so far.
Combining the peeling, change of measure, localization and concentration of measure steps
of the four previous sections, we have shown that
for all $\{\epsilon_{t,i,c}\}_{t,i}$, then
\beqan
\lefteqn{
[1] \eqdef
\Pr_{\theta^\star}\Big\{\bigcup_{1\leq n \leq t}\hat \theta_n\in\Theta_\rho\cap \cK_{a^\star}(\Pi_{a^\star}(\hat \nu_{a^\star,n}),\mu^\star-\epsilon) \geq f(t/n)/n\Big\}}\\
&\leq&
\sum_{c=1}^{C_{p,\eta,K}} \sum_{i=0}^{I_t-1}
\underbrace{
\exp\bigg(
- n_i\alpha^2
- \chi\sqrt{n_if(t/n_i)}\bigg)}_{\text{change of measure}}\bigg[
\underbrace{
	\exp\bigg(-\frac{n_i^2p\epsilon_{t,i,c}^2}{2V_{\rho} (n_{i+1}\!-\!1\!)}\bigg)\indic{\epsilon_{t,i,c}\leq \overline{\epsilon}_c }}_{\text{concentration}}\\
&&+
\underbrace{
\alpha_{p,K}\exp\Big(\!-\!f\Big(\frac{t}{n_{i+1}\!-\!1}\Big)\Big) \min\Big\{\rho^2v_\rho^2 , \epsilon_{t,i,c}^2,\frac{(K+2)v_\rho^2}{K (n_{i+1}\!-\!1\!)V_{\rho}}\Big\}^{-K/2}\epsilon_{t,i,c}^{K}
}_{\text{localization }+\text{ change of measure}}
\bigg]\,,
\eeqan
where we recall that $\alpha = \alpha(p,\eta,\epsilon) = \eta\rho_\epsilon\sqrt{v_\rho/2}$ and that the definition of $n_i$ is \beqan
n_i =
\begin{cases}
b^i  \text{ if } i < I_t \eqdef\lceil \log_b(\beta t+\beta)\rceil\\
t+1   \text{ if } i = I_t\,.
\end{cases}
\eeqan
A simple rewriting leads to the form
\beqan
[1] &\leq&
\sum_{c=1}^{C_{p,\eta,K}} \sum_{i=0}^{I_t-1}
\exp\bigg(
- n_i\alpha^2
- \chi\sqrt{n_if(t/n_i)}\bigg)\bigg[
\alpha_{p,K}\exp\Big(-f\Big(\frac{t}{n_{i+1}\!-\!1}\Big)\Big)\times\\
&&
\max\Big\{\frac{\epsilon_{t,i,c}}{\rho v_\rho}, 1,\sqrt{\frac{(\!n_{i+1}\!-\!1\!)V_\rho}{1\!+\!2/K}} \frac{\epsilon_{t,i,c}}{v_\rho}\Big\}^{K}+
\exp\bigg(-\frac{n_i^2p\epsilon_{t,i,c}^2}{2V_{\rho} (\!n_{i+1}\!-\!1\!)}\bigg)\indic{\epsilon_{t,i,c}\leq \overline{\epsilon}_c }
\bigg]\,,
\eeqan
which suggests we use $\epsilon_{t,i,c} = \sqrt{\frac{2V_{\rho} (\!n_{i+1}\!-\!1\!)f(t/(\!n_{i+1}\!-\!1\!))}{pn_i^2}}$.
Replacing this term in the above expression, we obtain
\beqan
[1] &\leq&\sum_{i=0}^{I_t-1}
\exp\bigg(
- n_i\alpha^2
- \chi\sqrt{n_if(t/n_i)}- f(t/(\!n_{i+1}\!-\!1\!))\bigg)f(t/(\!n_{i+1}\!-\!1\!))^{K/2}\times\\
&& C_{p,\eta,K}\Big(\alpha_{p,K}\max\Big\{\frac{2V_\rho}{p\rho^2 v^2_\rho b^{i-1}} , 1,\frac{b^2V_\rho^2}{pv_\rho^2(\frac{1}{2}\!+\!\frac{1}{K})} \Big\}^{K/2}
+1\Big)\,.
\eeqan
At this point, using the somewhat crude lower bound $b^i\geq 1$ it is convenient to introduce the constant
\beqan
C(K,\rho,p,b,\eta) =  C_{p,\eta,K}\Big(\alpha_{p,K}\max\Big\{\frac{2bV_\rho}{p\rho^2 v^2_\rho} , 1,\frac{b^2V_\rho^2}{pv_\rho^2(\frac{1}{2}\!+\!\frac{1}{K})} \Big\}^{K/2}
+1\Big)\,,
\eeqan
which leads to the final bound
\beqan
\lefteqn{
	\Pr_{\theta^\star}\Big\{\bigcup_{1\leq n \leq t}\hat \theta_n\in\Theta_\rho\cap \cK_{a^\star}(\Pi_{a^\star}(\hat \nu_{a^\star,n}),\mu^\star-\epsilon) \geq f(t/n)/n\Big\}}\\
&\leq&
C(K,\rho,p,b,\eta)
\sum_{i=0}^{I_t-1}
\exp\bigg(
- n_i\alpha^2
- \chi\sqrt{n_if(t/n_i)}- f(t/(\!n_{i+1}\!-\!1\!))\bigg)f(t/(\!n_{i+1}\!-\!1\!))^{K/2}\,.
\eeqan

\section{Fine-tuned upper bounds}\label{sec:tuning}
In this section, we study the behavior of the bound obtained in Theorem~\ref{thm:main} as a function of $t$, for a specific choice of function $f$, namely $f(x) = \log(x) + \xi\log\log x$, and prove corollary~\ref{cor:boundarycrossing} and
corollary~\ref{cor:boundarycrossingplus}, using a fine-tuning of the remaining free quantities.
This tuning is not completely trivial, as a naive tuning yields the condition that 
$\xi>K/2+1$ to ensure that the final bound is $o(1/t)$, while proceeding with some more care enables to show that $\xi>K/2-1$ is enough.
Let us remind that $f$ is non-decreasing only for $x\geq e^{-\xi}$.
We thus restrict to $t\geq e^{-\xi}$ in  corollary~\ref{cor:boundarycrossing} that uses the threshold $f(t)$, and to $\xi\geq 0$ in  corollary~\ref{cor:boundarycrossingplus} that uses the threshold function $f(t/n)$.
In the sequel, we use the short-hand notation $C$ 
in order to replace $C(K,\rho,p,b,\eta)$.

\subsection{Proof of Corollary~\ref{cor:boundarycrossing}}
As a warming-up, we start by the boundary crossing probability involving $f(t)$ instead of $f(t/n)$.
Indeed, controlling the boundary crossing probability
with term $f(t/n)$ is more challenging.
Although we focused so far on the boundary crossing probability with term $f(t/n)$, the previous proof directly applies to the case when $f(t)$ is considered.
In particular, the result of Theorem~\ref{thm:main}
holds also when all the terms $f(t/n),f(t/b^i), f(t/b^{i+1})$ are replaced with $f(t)$.

With the choice $f(x) = \log(x) + \xi\log\log x$, which is non-increasing on the set of $x$ such that $\xi>-\ln(x)$, Theorem~\ref{thm:main} specifies for all
$b>1,p,q,\eta\in(0,1)$,  
to
\beqan
\lefteqn{
\Pr_{\theta^\star}\Big\{\bigcup_{1\leq n < t}\hat \theta_n\in\Theta_\rho\cap \cK_{a^\star}(\Pi_{a^\star}(\hat \nu_{a^\star,n}),\mu^\star-\epsilon) \geq f(t)/n\Big\}}\\
&\leq& C\hspace{-5mm}\sum_{i=0}^{\lceil \log_b(qt)\rceil-1}
\hspace{-3mm}
\exp\bigg(
- \alpha^2b^i
- \chi\sqrt{b^if(t)}\bigg)e^{-f(t)}f(t)^{K/2}
\\
&=&\!\frac{C}{t}
\Bigg[\hspace{-1mm}\sum_{i=0}^{\lceil \log_b(qt)\rceil-1}
\hspace{-3mm}
\underbrace{
e^{
-\! \alpha^2 b^i
-\! \chi\sqrt{b^if(t)}}}_{s_i}\Bigg]
\!\log(t)^{K/2-\xi}
\bigg(1\! +\! \xi\frac{\log\log(t)}{\log(t)}\bigg)^{K/2}\,.
\eeqan

In order to study the sum $
S = \sum_{i=0}^{\lceil \log_b(qt)\rceil-1} s_i$ we provide two strategies. First, a direct upper bound gives
$S\leq \lceil \log_b(qt)\rceil
\leq \log_b(qt)+1$. Thus,
setting $q=1$ and $b=2$ we obtain
\beqan
\lefteqn{
\Pr_{\theta^\star}\Big\{\bigcup_{1\leq n < t}\hat \theta_n\in\Theta_\rho\cap \cK_{a^\star}(\Pi_{a^\star}(\hat \nu_{a^\star,n}),\mu^\star-\epsilon) \geq f(t)/n\Big\}}\\
&\leq&\frac{C}{t}
\bigg(1 + \underbrace{\xi\frac{\log\log(t)}{\log(t)}}_{o(1)}\bigg)^{K/2}\hspace{-2mm}\log(t)^{-\xi+K/2}(\log_2(t)+1)\,.
\eeqan
This term is thus $o(1/t)$ whenever $\xi>K/2+1$
and $O(1/t)$ when $\xi=K/2+1$.
We now show that a more careful analysis leads to a similar behavior even for smaller values of $\xi$. 
Indeed, let us note that for all $i\geq 0$, it holds by definition
\beqan
\frac{s_{i+1}}{s_i} &=&
\exp\bigg[ - \chi b^{i/2}(b^{1/2}-1)f(t)^{1/2} - \alpha^2b^i(b-1)\bigg]\\
&\leq&
\exp\bigg[ - \chi (b^{1/2}-1)f(t)^{1/2}\bigg]\,.
\eeqan
Since $f(t)\geq 1$, if we set $b = \lceil(1+\frac{\ln(1+\chi)}{\chi})^2\rceil$, which belongs to $(1,4]$ for all
$\chi\geq 0$, we obtain that $s_{i+1}/s_i \leq  \frac{1}{1+\chi}$.
Thus, we deduce that
\beqan
S \leq s_0 \sum_{i=0}^\infty(1+\chi)^{-i} = s_0 \frac{1+\chi}{\chi} = \frac{1+\chi}{\chi}\exp(-\alpha^2- \chi\sqrt{f(t)})\,.
\eeqan

Thus, $S$ is asymptotically $o(1)$, and we 
deduce that 
$
\Pr_{\theta^\star}\Big\{\bigcup_{1\leq n < t}\hat \theta_n\in\Theta_\rho\cap \cK_{a^\star}(\Pi_{a^\star}(\hat \nu_{a^\star,n}),\mu^\star-\epsilon) \geq f(t)/n\Big\}= o(1/t)$ beyond the condition $\xi>K/2+1$.
It is interesting to note that due to the term
$-\chi\sqrt{f(t)}$ in the exponent,
and owing to the fact that 
$\alpha\sqrt{\log(t)}-\beta\log\log(t) \to \infty$ for all positive $\alpha$ and all $\beta$,  we actually have the stronger property that $S\log(t)^{-\xi+K/2}= o(1)$ for all $\xi$
(using $\alpha=\chi$ and $\beta = K/2-\xi$).
However, since this asymptotic regime may take a massive amount of time to kick-in when $\alpha/\beta<1/2$ we do not advise to take $\xi$ smaller than $K/2-2\chi$. 
All in all, we obtain, for $C=  C(K,b,\rho,p,\eta)$ with
$b = \lceil(1+\frac{\ln(1+\chi)}{\chi})^2\rceil\leq 4$,
\beqan
\lefteqn{
	\Pr_{\theta^\star}\Big\{\bigcup_{1\leq n < t}\hat \theta_n\in\Theta_\rho\cap \cK_{a^\star}(\Pi_{a^\star}(\hat \nu_{a^\star,n}),\mu^\star-\epsilon) \geq f(t)/n\Big\}}\\
&\leq&\frac{C (1+\chi)}{t\chi}
\bigg(1 +\xi\frac{\log\log(t)}{\log(t)}\bigg)^{K/2}\hspace{-2mm}\log(t)^{-\xi+K/2}\exp(-\chi\sqrt{\log(t)+ \xi\log\log(t)})\,.
\eeqan

\subsection{Proof of Corollary~\ref{cor:boundarycrossingplus}}
Let us now focus on the proof of Corollary~\ref{cor:boundarycrossingplus} involving the  threshold $f(t/n)$.
We consider the choice $f(x) = \log(x) + \xi\log\log x$,
which is non-increasing on the set of $x$ such that $\xi>-\ln(x)$.
When $x=t/n$ and $n$ is about $t-O(\ln(t))$, ensuring this monotonicity property means that we require
$\xi$ to dominate $\ln(1- O(\ln(t)/t))$, that is $\xi\geq 0$.
Now, following the result of Theorem~\ref{thm:main},  we thus obtain for all
$b>1,p,q,\eta\in(0,1)$, 
\beqa
\lefteqn{
\Pr_{\theta^\star}\Big\{\bigcup_{1\leq n < t}\hat \theta_n\in\Theta_\rho\cap \cK_{a^\star}(\Pi_{a^\star}(\hat \nu_{a^\star,n}),\mu^\star-\epsilon) \geq f(t/n)/n\Big\}
}
\nonumber\\
&\!\leq\!&\!\! C \exp\bigg(
-\!\frac{\alpha^2q}{b}t-\! \chi\sqrt{\frac{tqf(b/q)}{b} }\bigg)+C\hspace{-5mm}\sum_{i=0}^{\lceil \log_b(qt)\rceil-2}
\!\!\!\hspace{-3mm}
\exp\!\bigg(\!\!
-\! \alpha^2b^i
-\! \chi\!\sqrt{b^if(t/b^i)}-\!f(t/(\!b^{i+1}\!-\!1\!))\bigg)
f\Big(\frac{t}{\!b^{i+1}\!-\!1\!}\Big)^{\!K/2}\nonumber\\
&\!=\!&\!\!
Ce^{-\!\frac{\alpha^2qt}{b}-\! \sqrt{\!\frac{\chi^2tq f(b/q)}{b}}\! }
\!+\!C\hspace{-5mm}\sum_{i=0}^{\lceil \log_b(qt)\rceil\!-\!2}
\hspace{-3mm}
\underbrace{
e^{
-\! \alpha^2b^i
-\! \chi\sqrt{b^if(t/b^i)}}
\Big(\frac{\!b^{i+1}\!\!-\!\!1\!}{t}\Big)
\!\log\Big(\frac{t}{\!b^{i+1}\!\!-\!\!1\!}\Big)^{\!K/2\!-\!\xi}}_{s_i}
\bigg(\!1\!\! +\!\! \underbrace{\xi\frac{\log\!\log\!\Big(\frac{t}{\!b^{i+1}\!-\!1\!}\Big)}{\log\!\Big(\!\frac{t}{(\!b^{i+1}\!-\!1\!)}\!\Big)}}_{o(1)}\bigg)^{\!K/2}\!\!\!\!\!\!\!
.\label{eqn:finedtuned}
\eeqa
We thus study the sum $
S = \sum_{i=0}^{\lceil \log_b(qt)\rceil-2} s_i$.
To this end, let us first study the term $s_i$.
Since $i\mapsto\log(t/b^{i+1})$ is a decreasing function of $i$, it holds for any index $i_0\in\Nat$ that
\beqan
s_i \leq 
\begin{cases}
	\Big(\frac{b^{i+1}}{t}\Big)
	\log\Big(\frac{t}{b\!-\!1}\Big)^{-\xi+K/2}&\text{if } \xi\leq K/2, i\leq i_0,\\
	\Big(\frac{b^{i+1}}{t}\Big)
	\log\Big(\frac{t}{b^{i_0+1}\!-\!1}\Big)^{-\xi+K/2}&\text{if } \xi\geq K/2, i\leq i_0,\\
	\exp(- \chi\sqrt{b^if(t/b^i)})\Big(\frac{b^{i+1}}{t}\Big)
\log\Big(\frac{t}{b^{i_0+1}\!-\!1}\Big)^{-\xi+K/2}&\text{if } \xi\leq K/2, i\geq i_0,\\
\exp(- \chi\sqrt{b^if(t/b^i)})\Big(\frac{b^{i+1}}{t}\Big)
\log\Big(\frac{1}{q}\Big)^{-\xi+K/2}&\text{if } \xi\geq K/2, i\geq i_0.
\end{cases}
\eeqan

{\bf Small values of $i$}
We start by handling the terms corresponding to small values of $i\leq i_0$, for some  $i_0$
to be chosen.
In that case, we note that
$r_i=\frac{b^{i+1}}{t}$ satisfies $r_{i-1}/r_i = 1/b<1$ and thus
\beqan
\sum_{i=0}^{i_0} s_i &\leq& s_{i_0}\sum_{i=0}^\infty (1/b)^i = \frac{bs_{i_0}}{b-1}\,,
\eeqan
from which we deduce that
\beqan
\sum_{i=0}^{i_0} s_i  \leq  \begin{cases}
 	\Big(\frac{bb^{i_0+1}}{t(b-1)}\Big)
 \log\Big(\frac{t}{b^{i_0+1}}\Big)^{K/2-\xi}&\text{if } \xi\geq K/2\\
 \Big(\frac{bb^{i_0+1}}{t(b-1)}\Big)
 \log\Big(\frac{t}{b\!-\!1}\Big)^{K/2-\xi}&\text{if } \xi\leq K/2\,.
 \end{cases} 
\eeqan
Following \cite{lai1988boundary}, in order to ensure that this quantity is summable in $t$, it is convenient to define $i_0$ as 
\beqan
i_0=\lfloor \log_b(t_0) \rfloor \quad\text{where}\quad
 t_0 = \frac{1}{c\log(ct)^{\eta}}\,,
\eeqan
for $\eta>K/2-\xi$ and a positive constant $c$.
Indeed in that case when $i_0\geq 0$ we obtain the bounds\footnote{This is also valid when $i_0<0$ since the sum is equal to $0$ in that case.}
\beqan
\sum_{i=0}^{i_0} s_i  \leq  \frac{b^2}{(b-1)ct \log(tc)^\eta}\times\begin{cases}
\log(tc\log(tc)^\eta/b)^{K/2-\xi}&\text{if } \xi\geq K/2\\
 \log(t/(b\!-\!1))^{K/2-\xi}&\text{if } \xi\leq K/2\,.
\end{cases} 
\eeqan
We easily see that this is $o(1/t)$ when both when $\xi>K/2$ 
and when $\xi\leq K/2$, by construction of $\eta$. Note that
$\eta$ can further be chosen to be equal to $0$ when  $\xi>K/2$.
The value of $c$ is  fixed by looking at what happens for larger values of $i\geq i_0$.
We note that the initial proof of \cite{lai1988boundary} uses the value $\eta=1$.

{\bf Large values of $i$}
We now consider the terms of the sum $S$ 
corresponding to large values $i>i_0$
 and thus focus on the term $s'_i = \exp(-\chi \sqrt{b^i \log(t/b^i)})b^{i+1}$, and better on the following ratio
   \beqan
\frac{s'_{i+1}}{s'_{i}} 
&=& b 
\exp\bigg[-\chi b^{i/2}\bigg(b^{1/2}\log\Big(\frac{t}{b^ib}\Big)^{1/2}
-\log\Big(\frac{t}{b^i}\Big)^{1/2}\bigg)\bigg]\,.
\eeqan
Remarking that this ratio is a non increasing function of $i$, we upper bound it by replacing $i$ with either $i_0+1$ or $0$. Using that $b^{i_0+1} \leq t_0$ we thus obtain,
\beqan
\frac{s'_{i+1}}{s'_{i}} 
&\leq&
\begin{cases} b 
	\exp\bigg[-\sqrt{\frac{\chi^2 }{c}}\bigg(
	\sqrt{
		\frac{b\log\Big(tc\log(tc)^\eta/ b\Big)}{\log(tc)^\eta}
	}
	-
	\sqrt{\frac{\log(tc\log(tc)^\eta) }{\log(tc)^\eta } }\bigg)
	\bigg]&\text{if }i_0\geq0\\
	b 
	\exp\bigg[-\chi \bigg(\sqrt{b\log\big(t/b\big)}
	-\sqrt{\log(t)}\bigg)\bigg]	& \text{else.}
\end{cases}
\eeqan
Since we would like this ratio to be less than $1$ for all (large enough) $t$, we readily see from this expression that this excludes the cases when $\eta>1$: the term in the ineer brackets converges to $0$ in such cases, and thus the ratio is asymptotically upper bounded by $b>1$.
Thus we impose that $\eta\leq 1$, that is $\xi \geq K/2-1$.

For the critical value $\eta=1$ it is then natural to study
the term $\sqrt{\frac{b\log(x\log(x)/b)}{\log(x)}}-\sqrt{\frac{\log(x\log(x))}{\log(x)}}$.
First, when $b=4$, this quantity is larger than $1/2$ for $x\geq 8.2$.
Then, it can be checked that $4\exp( - \frac{1}{2}\sqrt{\chi^2/c})<1$ if $c>\chi^2/(2\ln(4))^2$. These two conditions show that, provided that
$t \geq 8.2 (2\ln(4))^2\chi^{-2} \simeq 63 \chi^{-2}$, then $\frac{s'_{i+1}}{s'_{i}}<1$.
Now, in order to get a ratio $\frac{s'_{i+1}}{s'_{i}}$ that is away from $1$,
we target the bound $\frac{s'_{i+1}}{s'_{i}}<b/(b+1)$. This can be achieved by requiring that $t \geq 8.2 (2\ln(5))^2\chi^{-2} \simeq 85 \chi^{-2}$  by setting $c= \chi^2/(2\ln(5))^2$.
Eventually, we obtain for $b=4$  and $t \geq 85 \chi^{-2}$ the bound
\beqan
\sum_{i=i_0+1}^{I_t-2} s'_i &\leq& s'_{i_0+1}\sum_{i=i_0+1}^{I_t-2}(b/(b+1))^{i-i_0-1}
\leq s'_{i_0+1} (b+1)\\
& \leq &
(b+1)\exp\bigg[-\chi\sqrt{bt_0\log(t/bt_0) }\bigg] b^2t_0 \leq b^2(b+1)t_0\,.
\eeqan
\begin{remark}Another notable value is $\eta=0$.
	A similar study than the previous one shows that for $b=3.5$, 
the term $\sqrt{b\log(x/b)}-\sqrt{\log(x)}$ is larger than $1/2$ for $x>12$, which entails
that $\frac{s'_{i+1}}{s'_{i}}<b/(b+1)$ provided that  $t \geq 12 (2\ln(3.5))^2\chi^{-2} \simeq 76 \chi^{-2}$.
\end{remark}
 
Plugging-in the definition of $t_0$, and since $b^{i_0+1} \leq bt_0$, we obtain if $i_0\geq 0$, and for $b=4, c=\chi^2/(2\ln(5))^2$,
\beqa\label{eqn:largei}
\sum_{i=i_0+1}^{I_t-2} s_i \leq
\begin{cases}\frac{b^2(b+1)}{tc\log(tc)} \log(1/q)^{K/2-\xi}
	&\text{if } \xi\geq K/2\\
\frac{b^2(b+1)}{tc\log(tc)} \log(t\frac{c\log(tc)}{b-c\log(tc)})^{K/2-\xi}
&\text{if } \xi\in[K/2-1,K/2]\,.
\end{cases}
\eeqa

It remains to handle the case when $i_0<0$.
Note that this case only happens for $t$ large enough so that $t>c^{-1}e^{\frac{1}{bc}}$. The later quantity may be huge since $1/bc = \log(5)^2\chi^{-2}$ is possibly large when $\chi$ is close to $0$.
 In that case, we directly control $\sum_{i=0}^{I_t-2}s_i$. 
We control the ratio $s'_{i+1}/s'_i$ by $b/(b+1/2)$ provided that
\beqan
\sqrt{b\log(t/b)}- \sqrt{\log(t)}>\frac{\log(b+1/2)}{\chi}, \text{where } b=4\,.
\eeqan
Thus, if we define $t_\chi$ to be the smallest such $t$, then when
 $t>c^{-1}e^{\frac{1}{bc}}$ and provided that $t\geq t_\chi$,
 the bound of \eqref{eqn:largei} remains valid for the sum $S$, up to replacing $b^2(b+1)$
 with $2b^2(b+1/2)$ and 
 $\log(t\frac{c\log(tc)}{b-c\log(tc)})$ with 
$\log(t/(b-1))$.
  The later constraint  $t\geq t_\chi$ is satisfied as soon as $4\ln(5)^2\chi^{-2}e^{\chi^{-2}\ln(5)^2}\geq t_\chi$  which is generally satisfied for $\chi$ not too large.

\paragraph{Final control on S}
We can now control the term $S$ by combining the two bounds for large and small $i$.
We get for $c = \chi^{2}/(2\ln(4.5))^2$ and $b=4$,
and provided that  $t \geq 85\chi^{-2}$
 and $t\leq  \chi^{-2}\frac{\exp\big(\chi^{-2}\ln(4.5)^2\big)}{4\ln(4.5)^2}$, the following bound
\beqa\label{eqn:boundonS}
S \leq
\frac{b}{ct \log(tc)}
\begin{cases}
	 \frac{b}{(b-1)}\log(tc\log(tc)/b)^{K/2-\xi}
	+b(b+1) \log(1/q)^{K/2-\xi}&\text{if } \xi\geq K/2\\
	 \frac{b}{(b-1)}\log(t/(b-1))^{K/2-\xi}+ b(b+1) \log(t\frac{c\log(tc)}{b-c\log(tc)})^{K/2-\xi}&\text{if }  \xi\in[K/2-1,K/2]\,.
\end{cases} 
\eeqa

Further, for larger values of $t$, $t\geq \chi^{-2}\frac{\exp\big(\chi^{-2}\ln(4.5)^2\big)}{4\ln(4.5)^2}$, then
\beqa\label{eqn:boundonS2}
S \leq
\frac{2b^2(b+1/2) }{ct \log(tc)}
\begin{cases}
\log(1/q)^{K/2-\xi}&\text{if } \xi\geq K/2\\
\log(t/(b-1))^{K/2-\xi}&\text{if }  \xi\in[K/2-1,K/2]\,.
\end{cases} 
\eeqa

\paragraph{Concluding step}
In this final step, we now gather equation \eqref{eqn:finedtuned}
together with the previous bounds \eqref{eqn:boundonS},  \eqref{eqn:boundonS2} on $S$. 
We obtain that for all $p,q,\eta\in(0,1)$
\beqan
\lefteqn{
	\Pr_{\theta^\star}\Big\{\bigcup_{1\leq n < t}\hat \theta_n\in\Theta_\rho\cap \cK_{a^\star}(\Pi_{a^\star}(\hat \nu_{a^\star,n}),\mu^\star-\epsilon) \geq f(t/n)/n\Big\}}\\
& \leq  &
C(K,\rho,p,b,\eta)\bigg(e^{-\frac{\alpha^2qt}{b}- \sqrt{\frac{\chi^2tq f(b/q)}{b}} }
 +\!S(1 +\xi)^{K/2}\bigg)\,.
\eeqan
 where we re call the definition of the constants
$\alpha = \eta \rho_\epsilon\sqrt{v_\rho/2},\quad \chi = p\eta\rho_\epsilon \sqrt{2v_\rho^2/V_\rho}.$

When $\xi \in [K/2-1,K/2]$, one can then choose $q=1$.
When $\xi\geq K/2$, there is a trade-off in $q$, since the first term (the exponential)
is decreasing with $q$ while the  second term is increasing with $q$.
For instance choosing $q = \exp(- \kappa^{-1/\eta})$, where $\eta = \xi-K/2$ and $\kappa>0$ leads to  $\log(1/q)^{K/2-\xi} = \kappa$. 
When $b=4$, simply  choosing $q=0.8$ gives the final bound after some cosmetic simplifications.

\section*{Conclusion}
In this work, that should be considered as a tribute to the contributions of  T.L. Lai, we shed light on a beautiful and seemingly forgotten result from \cite{lai1988boundary}, that we modernized into a fully non-asymptotic statement,
with explicit constants that can be directly used, for instance, for the regret analysis of multi-armed bandits strategies.
Interestingly, the final results, whose roots are thirty-years old, 
show that the existing analysis of \KLUCB\ that was only stated for exponential families of dimension $1$ and discrete distributions lead to a sub-optimal constraints on the tuning of the threshold function $f$,
and can be extended to work with exponential families of arbitrary dimension $K$ and even for the thresholding term of the \KLUCBp\ strategy, whose analysis was left open.

This proof technique is mostly based on a change-of-measure argument, like the lower bounds for the analysis of sequential decision making strategies and in stark contrast with other key results in the literature (\cite{HondaTakemura10DMED, MaillardMunosStoltz11klucb, CaGaMaMuSt2013}).
We believe  and hope that the novel writing of this proof technique that we provided here will greatly benefit  the community working on boundary crossing probabilities, sequential design of experiments as well as stochastic decision making strategies.


\newpage
\appendix

\section{Technical details}\label{app}

\paragraph{Lemma~\ref{lem:dim1KL} (Dimension 1)}
\emph{
	Consider a canonical one-dimensional family (that is $K=1$ and  $F_1(x)=x\in\Real$). Then, for all $f$ such that $f(t/n)/n$ is non-increasing in $n$, then
	\beqan
	\Pr_{\theta^\star}\Big\{\bigcup_{m\leq n < M} \,\, \cB^\psi(\hat \theta_n,\theta^\star) \geq f(t/n)/n \Big\}&\leq&\exp\bigg(-\frac{m}{M}f(t/M)\bigg)\,.
	\eeqan
}

\begin{myproof}{of Lemma~\ref{lem:dim1KL}}
	The proof goes as follows. First, we observe that:
	\beqan
	\Pr_{\theta^\star}\Big\{\bigcup_{m\leq n < M} \,\, \cB^\psi(\hat \theta_n,\theta^\star) \geq f(t/n)/n \Big\}&=&
	\Pr_{\theta^\star}\Big\{\bigcup_{m\leq n < M} \,\, \Phi^\star(\hat F_n) \geq f(t/n)/n \Big\}\\
	&\leq&
	\Pr_{\theta^\star}\Big\{\bigcup_{m\leq n < M} \,\, \Phi^\star(\hat F_n) \geq f(t/M)/M \Big\}\,.
	\eeqan
	At this point note that if for all $F=\nabla \psi(\theta)$  with mean $\mu_\theta\leq \mu^\star-\epsilon$, it holds that $\Phi^\star(F)<f(t/M)/M$
	then the probability of interest is $0$ and we are done.
	In the other case, there exists an $F_M$ such that $\Phi^\star(F_M)=f(t/M)/M$.  We thus proceed with this case as follows
	\beqan
	\lefteqn{
		\Pr_{\theta^\star}\Big\{\bigcup_{m\leq n < M} \,\, \cB^\psi(\hat \theta_n,\theta^\star) \geq f(t/n)/n \Big\}}\\
	&\leq&\Pr_{\theta^\star}\Big\{\bigcup_{m\leq n < M} \,\, \Phi^\star(\hat F_n) \geq \Phi^\star(F_M) \Big\}\\
	&\stackrel{(a)}{\leq}& 
	\Pr_{\theta^\star}\Big\{\bigcup_{m\leq n < M} \,\, \hat F_n \leq F_M \Big\}\\
	&\stackrel{(b)}{\leq}& 
	\Pr_{\theta^\star}\Big\{\bigcup_{m\leq n < M} \,\, \exp\bigg(\lambda \sum_{i=1}^nF(X_i)\bigg)\geq \exp\bigg(n\lambda F_M\bigg) \Big\}\\
	&\leq&
	\Pr_{\theta^\star}\Big\{\bigcup_{m\leq n < M} \,\, \exp\bigg(\sum_{i=1}^n\Big(\lambda F(X_i)-\Phi(\lambda)\Big)\bigg)\geq \exp\bigg(n [\lambda F_M - \Phi(\lambda)]\bigg) \Big\}\\
	&\stackrel{(c)}{\leq}& 
	\Pr_{\theta^\star}\Big\{\max_{m\leq n < M} \,\, \exp\bigg( \sum_{i=1}^n\Big(\lambda F(X_i)-\Phi(\lambda)\Big)\bigg)\geq \exp\bigg(m [\lambda F_M - \Phi(\lambda)]\bigg) \Big\}\,,
	\eeqan
	where (a) holds by \eqref{eqn:PhistarDec},
	(b) holds for all $\lambda<0$, and (c) for all
	$\lambda<0$ such that $\lambda F_M - \Phi(\lambda)>0$. Now,  the process defined by 
	$W_{\lambda,0}=1$ and $W_{\lambda,n}=\exp\bigg(\sum_{i=1}^n\Big( \lambda F(X_i)-\Phi(\lambda)\Big)\bigg)$ is a non-negative super-martingale, since it holds
	\beqan
	\Esp_{\theta^\star}\bigg[\exp\bigg(\sum_{i=1}^n\Big(\lambda F(X_i)-\Phi(\lambda)\Big)\bigg)\bigg|\cH_{n-1}\bigg]
	&=&
	W_{\lambda,n-1}
	\Esp_{\theta^\star}\bigg[\exp\bigg(\lambda F(X_n)-\Phi(\lambda)\bigg)\bigg|\cH_{n-1}\bigg]\\
	&\leq&
	W_{\lambda,n-1}
	\exp\bigg(\Phi(\lambda)-\Phi(\lambda)\bigg) \leq 1\,.
	\eeqan
	Thus, we deduce that for all $\lambda<0$ such that $\lambda F_M - \Phi(\lambda)>0$
	\beqan
	\Pr_{\theta^\star}\Big\{\bigcup_{m\leq n < M} \,\, \cB^\psi(\hat \theta_n,\theta^\star) \geq f(t/n)/n \Big\}&\leq&\exp\bigg(-m [\lambda F_M - \Phi(\lambda)]\bigg)\,.
	\eeqan
	Since by \eqref{eqn:Supremum} this is satisfied by the optimal $\lambda$ for $\Phi^\star(F_M)$, we thus deduce that
	\beqan
	\Pr_{\theta^\star}\Big\{\bigcup_{m\leq n < M} \,\, \cB^\psi(\hat \theta_n,\theta^\star) \geq f(t/n)/n \Big\}&\leq&\exp\bigg(-m \Phi^\star(F_M)\bigg)=\exp\bigg(-\frac{m}{M}f(t/M)\bigg)\,.
	\eeqan
	
	\vspace{-6mm}
\end{myproof}

\paragraph{Lemma~\ref{lem:volumecones}}
\emph{
	For all $\epsilon,\epsilon'>0$, $p,p'\in[0,1]$ and all $K\geq 1$ the following equality holds
	\beqan
	\frac{
		|\cB_2(0,\epsilon) \cap \cC_p(0;{\bf 1})|}{\int_{\cB_2(0,\epsilon') \cap \cC_{p'}(0;{\bf 1})}e^{- \norm{y}^2/2} dy }&=& \frac{\omega_{p,K-2}}{\omega_{p',K-2}}\frac{\int_{0}^\epsilon r^{K-1} dr}{\int_{0}^{\epsilon'} e^{-r^2/2} r^{K-1}dr}\,,
	\eeqan
	where  $\omega_{p,K-2}=\int_{p}^1 \sqrt{1-z^2}^{K-2}dz$ for $K\geq 2$ and  
	using the convention that $\omega_{p,-1}=1$.
	Further, 
	\beqan
	\frac{
		|\cB_2(0,\epsilon) \cap \cC_p(0;{\bf 1})|}{\int_{\cB_2(0,\epsilon') \cap \cC_{p'}(0;{\bf 1})}e^{- \norm{y}^2/2} dy }&\leq& 
	2\frac{\omega_{p,K-2}}{\omega_{p',K-2}} \bigg(\frac{\epsilon}{ \min \{ \epsilon',\sqrt{1+2/K}\}}\bigg)^K\,.
	\eeqan}
\begin{myproof}{of Lemma~\ref{lem:volumecones}}
	First of all, let us remark that  provided that $K\geq 2$, then
	\beqan
	|\cB_2(0,\epsilon) \cap \cC_p(0;{\bf 1})| &=& \int_{0}^\epsilon
	|\{y \in\Real^K : \langle y, {\bf 1}\rangle \geq r p, \norm{y} = r  \}| dr\\
	&=& \int_{0}^\epsilon \int_{rp}^r 
	|\{y \in\Real^K :y_1=z, \norm{y} = r  \}| dzdr\\
	&=& \int_{0}^\epsilon \int_{rp}^r 
	|\{y \in\Real^{K-1} :\norm{y} = \sqrt{r^2-z^2}  \}| dzdr\\
	&=& \int_{0}^\epsilon r^{K-1}\int_{p}^1 \sqrt{1-z^2}^{K-2}|\cS_{K-1}| dzdr\,.
	\eeqan
	where $\cS_{K-1}\subset\Real^{K-1}$ is the $K-2$ dimensional unit sphere of $\Real^{K-1}$.  Let us recall that when $K=2$, we get $|\cS_{K-1}|=2$.
	For convenience, let us denote $\omega_{p,K-2}=\int_{p}^1 \sqrt{1-z^2}^{K-2}dz$.
	Then, for $K\geq 2$,
	\beqan
	|\cB_2(0,\epsilon) \cap \cC_p(0;{\bf 1})| &=& |\cS_{K-1}|  \int_{0}^\epsilon r^{K-1} \omega_{p,K-2}dr\,.
	\eeqan	
	For $K=1$,  $|\cB_2(0,\epsilon) \cap \cC_p(0;{\bf 1})|=\epsilon$.
	Likewise, we obtain, following the same steps that
	\beqan
	\int_{\cB_2(0,\epsilon) \cap \cC_p(0;{\bf 1})}e^{- \norm{y}^2/2} dy &=&
	|\cS_{K-1}| \int_{0}^\epsilon e^{-r^2/2} r^{K-1} \omega_{p,K-2}dr\,.
	\eeqan
	
	We obtain the first part of the lemma by combining the two previous equalities.
	For the second part, we use the inequality $e^{-x} \geq 1-x$, which gives
	\beqan
	\int_{0}^\epsilon e^{-r^2/2} r^{K-1} dr \geq  \int_{0}^\epsilon r^{K-1} - \frac{1}{2}r^{K+1} dr = \epsilon^K\Big(\frac{1}{K}- \frac{\epsilon^{2}}{2(K+2)}\Big)\,.
	\eeqan
	Thus, whenever $\epsilon^2<(K+2)/K$, we obtain
	\beqan
	\int_{0}^\epsilon e^{-r^2/2} r^{K-1} dr  \geq  \frac{\epsilon^K}{2K}.
	\eeqan
	On the other hand, if $\epsilon^2\geq (K+2)/K$, then
	\beqan
	\int_{0}^\epsilon e^{-r^2/2} r^{K-1} dr &\geq& \int_{0}^{(K+2)/K} e^{-r^2/2} r^{K-1} dr\\
	&\geq&\frac{\sqrt{1+2/K}^K}{2K}\,.
	\eeqan
	Thus, in all cases, the integral is larger than $\frac{\min\{\epsilon, \sqrt{1+2/K}\}^K}{2K}$, and we conclude by simple algebra.	
\end{myproof}

\end{document}